\documentclass[journal]{IEEEtran}
\ifCLASSINFOpdf
\else
\fi
\usepackage{amsmath,amsfonts,bm}

\def\eqref#1{equation~\ref{#1}}

\def\1{\bm{1}}

\def\mA{{\bm{A}}}
\def\mB{{\bm{B}}}

\def\mD{{\bm{D}}}

\def\mL{{\bm{L}}}

\def\mP{{\bm{P}}}

\def\mR{{\bm{R}}}
\def\mS{{\bm{S}}}

\DeclareMathAlphabet{\mathsfit}{\encodingdefault}{\sfdefault}{m}{sl}
\SetMathAlphabet{\mathsfit}{bold}{\encodingdefault}{\sfdefault}{bx}{n}

\usepackage{hyperref}
\usepackage{url}

\usepackage{booktabs}                                   
\usepackage{multirow}                                   
\usepackage{tablefootnote}                              
\usepackage[symbol]{footmisc}                           
\usepackage{amsmath,amssymb}                            
\usepackage{xcolor}                                     
\usepackage{enumitem}                                   
\usepackage{subcaption}                                 
\usepackage{graphicx}
\usepackage{arydshln}
\usepackage{cleveref}

\usepackage{algorithm}
\usepackage{algpseudocode}
\usepackage{float}
\usepackage{wrapfig}

\newcommand{\eat}[1]{}                                  

\usepackage{cite}
\usepackage{booktabs}               
\usepackage{nicematrix}
\usepackage{subcaption}
\usepackage{xcolor}
\usepackage{wrapfig}

\usepackage{amsmath}
\usepackage{amsthm}

\usepackage{bm}

\usepackage{amssymb}
\usepackage[T1]{fontenc}

\usepackage{pifont}
\newtheorem{theorem}{Theorem}[section]
\newtheorem{prop}{Proposition}[section]
\newtheorem{defi}{Definition}[section]

\newtheorem{lemma}[theorem]{Lemma}

\allowdisplaybreaks

\newcommand{\maL}        {\mathcal{L}}

\newcommand{\bbE}     {\mathbb{E}}
\newcommand{\bbR}     {\mathbb{R}}

\newcommand{\nexta}     {a'}
\newcommand{\nexts}     {s'}
\newcommand{\hatQ}      {\hat{Q}}
\newcommand{\hatV}      {\hat{V}}
\newcommand{\hatP}      {\hat{\pi}}
\newcommand{\hatB}      {\hat{\mathcal{B}}}

\newcommand{\rmarg}     {{\rm arg}}
\newcommand{\rmmin}     {\mathop{{\rm min}}}
\newcommand{\rmmax}     {\mathop{{\rm max}}}
\newcommand{\rmsupp}     {\mathop{{\rm supp}}}
\newcommand{\BH}        {\pi_{\beta}}

\newcommand{\dif}[1]{#1}

\begin{document}

\title{ACL-QL: Adaptive Conservative Level in Q-Learning for Offline Reinforcement Learning}

\author{Kun Wu,~\IEEEmembership{Student Member,~IEEE},
        Yinuo Zhao,
        Zhiyuan Xu,~\IEEEmembership{Member,~IEEE}, 
        Zhengping Che,~\IEEEmembership{Member,~IEEE},
        Chengxiang Yin,
        Chi Harold Liu,~\IEEEmembership{Senior Member,~IEEE},
        Feifei Feng,
        and Jian~Tang,~\IEEEmembership{Fellow,~IEEE}

\thanks{Kun Wu and Chengxiang Yin are with the Department of Electrical Engineering and Computer Science,
Syracuse University, Syracuse, NY, 13244 USE. E-mail: \{kwu102, cyin02\}@syr.edu}
\thanks{Yinuo Zhao and Chi Harold Liu are with Beijing Institute of Technology, Beijing 100811, China. E-mail: ynzhao@bit.edu.cn, liuchi02@gmail.com} 
\thanks{Zhiyuan Xu, Zhengping Che, and Jian Tang are with Beijing Innovation Center of Humanoid Robotics, Beijing 101111, China. E-mail: \{Eric.Xu, z.che, jian.tang\}@x-humanoid.com, chezhengping@gmail.com} 
\thanks{Feifei Feng is with Midea Group, Shanghai201700, China. E-mail: feifei.feng@midea.com} 
\thanks{Corresponding author: Jian Tang}
\thanks{This article has supplementary downloadable material available at https://doi.org/10.1109/TNNLS.2024.3497667, provided by the authors.}
\thanks{Digital Object Identifier 10.1109/TNNLS.2024.3497667}
}

\maketitle

\begin{abstract}
    Offline Reinforcement Learning (RL), which operates solely on static datasets without further interactions with the environment, provides an appealing alternative to learning a safe and promising control policy.
The prevailing methods typically learn a conservative policy to mitigate the problem of Q-value overestimation, but it is prone to overdo it, leading to an overly conservative policy.
Moreover, they optimize all samples equally with fixed constraints, lacking the nuanced ability to control conservative levels in a fine-grained manner.
Consequently, this limitation results in a performance decline.
%
To address the above two challenges in a united way, we propose a framework, Adaptive Conservative Level in Q-Learning (ACL-QL), which limits the Q-values in a mild range and enables adaptive control on the conservative level over each state-action pair, i.e., lifting the Q-values more for good transitions and less for bad transitions.
We theoretically analyze the conditions under which the conservative level of the learned Q-function can be limited in a mild range and how to optimize each transition adaptively.
Motivated by the theoretical analysis, we propose a novel algorithm, ACL-QL, which uses two learnable adaptive weight functions to control the conservative level over each transition.
Subsequently, we design a monotonicity loss and surrogate losses to train the adaptive weight functions, Q-function, and policy network alternatively.
We evaluate ACL-QL on the commonly used D4RL benchmark and conduct extensive ablation studies to illustrate the effectiveness and state-of-the-art performance compared to existing offline DRL baselines.
\end{abstract}


\begin{IEEEkeywords}
Model-free reinforcement learning (RL), offline RL, RL
\end{IEEEkeywords}

%
\IEEEpeerreviewmaketitle


\section{Introduction}

With the help of deep learning, Reinforcement Learning (RL) has achieved remarkable results on a variety of previously intractable problems, such as playing video games~\cite{Silver2016mastering,Ren2018self,Dilokthanakul2019feature}, controlling robot~\cite{Kalashnikov2018scalable,Akkaya2019solving, Garaffa2021reinforcement} and driving autonomous cars~\cite{Huang2021deductive,Wu2021deep,Zhao2022cadre}.
However, the prerequisite that the agent has to interact with the environments makes the learning process costly and unsafe for many real-world scenarios.
Recently, offline RL~\cite{Lange2012batch,Prudencio2023survey} has been proposed as a promising alternative to relax this requirement. 
In offline RL, the agent directly learns a control policy from a given static dataset, which is previously collected by an unknown behavioral policy. 
Offline RL enables the agent to achieve comparable or even better performance without additional interactions with the environment. 
~\cite{Prudencio2023survey} provides a comprehensive review of the offline RL problem.

Unfortunately, stripping the interactions from the online RL, offline RL is very challenging due to the distribution shift between the behavioral policy and the learned policy over the training process.
It often leads to the overestimation of values of out-of-distribution (OOD) actions~\cite{Kumar2019stabilizing,Levine2020offline} and thus misleads the policy into choosing these erroneously estimated actions.
While the target Q-values are calculated using OOD actions and have erroneously high values, the policy is then optimized towards predicting OOD actions. 
This error then propagates and intensifies with bootstrapping during training, eventually leading to the explosion and overflow of the Q-values.
In online RL, the interactions with the environment could provide the right feedback and thus fix such errors, yet it is impossible for offline RL.
To alleviate the values overestimation problem, recent methods~\cite{Kumar2019stabilizing,Jaques2019way,Wu2019behavior,Siegel2020keep} proposed to constrain the learned policy to the behavioral policy in different ways, such as limiting the action space~\cite{Fujimoto2019off}, using KL divergence~\cite{Wu2019behavior}, and using Maximum Mean Discrepancy (MMD)~\cite{Kumar2019stabilizing}.
Besides directly constraining the policy, other methods~\cite{Kumar2020conservative,Yu2021conservative,Yu2021combo,Ma2021conservative} choose to learn a conservative Q-function to constrain the policy implicitly and thus alleviate the overestimation problem of the Q-function. 
However, such value-constrained methods often suffer from the over-conservative problem~\cite{Kumar2020conservative, Yu2021combo} that suppresses the Q-values of OOD actions too much to learn a satisfactory policy.
Having observed the above phenomenon, we try to find a moderate range of conservatism that allows the Q-function to avoid the problem of overestimation while avoiding the problem of being too conservative.
Specifically, we derive the conditions on how to limit the learned Q-function lies between the ordinary Q-function and the over-conservative Q-function in Conservative Q-Learning (CQL)~\cite{Kumar2020conservative}, and subsequently optimize the Q-function based on the conditions.

\begin{figure}
    \begin{center}
    \includegraphics[width=0.5\textwidth]{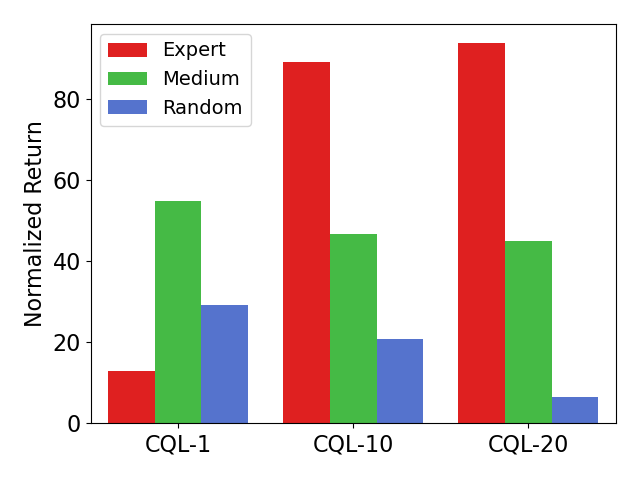}
    \end{center}
    \caption{Performance gaps of CQL~\cite{Kumar2020conservative} with different conservative levels ($\alpha=1,10,20$) on HalfCheetah-v2 tasks with expert, medium, random datasets.}
    \label{fig:intro_cql}
\end{figure}

In addition, most previous methods meet the following two challenges:
(1) They usually require prior knowledge (e.g., the quality of each transition) and expert experience to set suitable hyperparameters and thus increase the trial-and-error costs for deployment.
(2) Even given the suitable hyperparameters, such as a fixed threshold value to constrain the distribution gap between the learned policy and the behavioral policy, it is still difficult to learn an exceptional policy because all transition samples are optimized equally with the fixed threshold rather than controlling the conservative levels of all samples in a more fine-grained way.
An example of the seminal work CQL~\cite{Kumar2020conservative} showing the above two challenges is provided in Figure~\ref{fig:intro_cql}.
CQL uses a hyperparameter $\alpha$ to control the conservative level on all transition samples. 
A higher $\alpha$ represents CQL squeezes Q-values of OOD actions down more and raises Q-values of in-dataset actions up.
A more conservative (e.g., $\alpha=20$) CQL agent achieves higher returns on the expert dataset while suffering from performance degradation on the random dataset, indicating for high-quality data, a higher conservative level works better and vice versa. 
It clearly shows the significant influence of different conservative levels on the final results for varied dataset types.
Therefore, it is more proper to use adaptive weights for different transition samples to control the conservative level in a fine-grained way, such as raising the Q-values more for good transitions and less for bad transitions.
The adaptive weights can provide at least the following benefits to the above problems:
(1) Compared to using fixed parameters to constrain all data, fine-grained control for each transition sample can help learn a more accurate Q-function and better policy. 
(2) Compared to the need for prior knowledge of the dataset, we only need to specify an approximate range of the conservative level, and the fine-grained conservative levels will be automatically learned within this range, thus reducing the difficulty and costs for deployment.

In this paper, we focus on how to constrain the Q-function in a mild range and adaptively learn suitable conservative levels over each transition.
Therefore, we propose a general framework named Adaptive Conservative Level in Q-Learning (ACL-QL), which sheds light on how to design a proper adaptive Q-function.
In ACL-QL, we present two adaptive weight functions to estimate the conservative level weights for each in-dataset and OOD state-action pair, respectively.
To mitigate the overestimation and over-conservative problem in Q-function simultaneously, we take the Q-values of the ordinary Q-function and overly conservative Q-function in CQL~\cite{Kumar2020conservative} as anchors (i.e., one as an upper bound and one as a lower bound), thus forming a moderate range.
We theoretically discuss in detail the correlation between the different conservative levels and their corresponding conditions that the weight functions need to be satisfied.
To achieve more fine-grained control over the conservative level of the Q-function, the form of the adaptive weight functions is not fixed, and we are able to define particular forms according to practical needs, such as neural networks or specified rules.
Therefore, the adaptive weight functions can output different weights for every state-action pair.
We also formally define the monotonicity of the weight functions to depict the property that weight functions should raise the Q-values more for good transitions and less for bad transitions.

With the guidance of theoretical conditions, we propose one practical algorithm named ACL-QL with learnable neural networks as adaptive weight functions.
Overall, ACL-QL consists of three steps. 
Firstly, we preprocess the static dataset to calculate the relative transition quality measurements and use them as pseudo labels to learn the adaptive weights.
Then, with the help of the measurements, we construct a monotonicity loss to maintain the monotonicity of adaptive weight functions with respect to the transition quality measurements.
We also convert previous theoretical conditions into surrogate losses to keep the conservative level of ACL-QL in a mild range, which is between the ordinary Q-function and the overly conservative Q-function in CQL.
Lastly, we train the adaptive weight network, actor network, and critic network alternatively.

We summarize our contributions as follows:
1) We propose a framework named Adaptive Conservative Level in Q-Learning (ACL-QL) that supports the fine-grained control of the conservative level in offline DRL.
%
2) We theoretically analyze how the conservative level changes conditioned on different forms of adaptive weight functions. 
%
3) With the guidance of the proposed framework, we present a novel practical algorithm  ACL-QL with carefully designed surrogate and monotonicity losses to control the conservative levels and monotonicity. 
4) We conducted extensive experiments on the D4RL benchmark, and the state-of-the-art results demonstrated the effectiveness of our framework.

\section{Related Work}

\textbf{Imitation Learning.}
To learn from a given static dataset, Imitation Learning (IL) is the most straightforward strategy.
The core spirit of IL is to mimic the behavioral policy.
As the simplest form, behavioral cloning still holds a place for offline reinforcement learning, especially for the expert dataset.
However, having expert datasets is only a minority of cases.
Recently, some methods~\cite{Chen2020bail,Siegel2020keep,Wang2020critic,Liu2021curriculum} aim to filter sub-optimal data and then apply the supervised learning paradigm afterward. 
Specifically, Best-Action Imitation Learning (BAIL)~\cite{Chen2020bail} performed imitation learning only on a high-quality subset of the dataset purified by a learned value function.
However, these methods often neglect the information contained in the bad actions with lower returns and thus often fail in tasks with non-optimal datasets.
We believe these methods are beneficial to our ACL-QL framework since they split different data regions where different conservative levels can be set.

\textbf{Off-policy RL.}
Off-policy RL methods aim to improve the sample efficiency compared to on-policy RL methods in the online RL setting but face a variety of challenges, including overestimation bias~\cite{Thrun2014issues,Fujimoto2019off}, instability~\cite{Fu2019diagnosing}, divergence~\cite{Achiam2019towards} and delusional bias~\cite{Lu2018non}.
Soft Actor-Critic (SAC)~\cite{Haarnoja2018soft} and Twin Delayed Deep Deterministic policy gradient (TD3)~\cite{Fujimoto2018addressing} use a replay buffer to collect previous transitions for optimizing the actor.
Adaptive Q-Learning~\cite{Luo2018adaptive} is an online RL method that achieves an adaptive balance between the current and future Q-functions. 
Prioritized Experience Replay (PER)~\cite{Schaul2015prioritized} is one seminal work incorporated into the Deep Q-Networks (DQN) algorithm~\cite{Mnih2013playing}.
Subsequently, a lot of modifications and variants have been proposed, including Distributed PER~\cite{Horgan2018distributed}, Loss Adjusted Prioritized (LAP) Experience Replay~\cite{Fujimoto2020equivalence}, and Model-Augmented PER (MaPER)~\cite{Oh2021model}.
However, directly applying the off-policy RL methods to the offline RL problem still suffers from the distributed shift and value overestimation problems~\cite{Lange2012batch,Prudencio2023survey}.
In contrast, ACL-QL uses learnable weight functions to enable adaptive conservative levels in Q-Learning, which is specifically designed for the offline RL problem setting.
Other methods focus on applying RL methods to time-varying systems~\cite{Hamadanian2022demystifying,Khetarpal2022towards}, where an exogenous factor changes the dynamics of the system over time, and the challenges lie in catastrophic forgetting~\cite{Atkinson2021pseudo} and exploration~\cite{Ahmed2019understanding}. 
In this work, we aim to mitigate the distribution shift and value overestimation problems presented in the offline RL setting.

\textbf{Model-free Offline RL.}
A large number of model-free offline RL~\cite{Fujimoto2019off,Kumar2020conservative,Kostrikov2021implicit,Bai2022monotonic,Huang2023mild} methods aim to maximize the returns while constraining the learned policy and the behavioral policy to be close enough.
There are various ways for the direct constraint on policy, including minimizing the KL-divergence~\cite{Jaques2019way,Wu2019behavior,Zhou2020plas},
MMD~\cite{Kumar2019stabilizing}, Fisher-divergence~\cite{Kostrikov2021offline} or Wasserstein distance~\cite{Wu2019behavior}, and adding behavioral cloning regularization~\cite{Fujimoto2021minimalist}.
The policy can also be constrained implicitly by actions space reduction~\cite{Fujimoto2019off},
importance sampling based algorithms~\cite{Sutton2016emphatic,Nachum2019algaedice}, 
the implicit form of KL-divergence~\cite{Nair2020accelerating,Peng2019advantage,Simao2020safe},%
uncertainty quantification~\cite{Agarwal2020optimistic,Kumar2019stabilizing} or a conservative Q-function~\cite{Kumar2020conservative,Ma2021conservative,Sinha2022s4rl,Yu2021conservative}. 
More recently, Onestep RL~\cite{Brandfonbrener2021offline} and Implicit Q-Learning (IQL)~\cite{Kostrikov2021implicit} proposed to improve the policy after the convergence of Q functions. 
Trajectory Transformer (TT)~\cite{Janner2021offline} and Decision Transformer (DT)~\cite{Chen2021decision} leverages the advantage of Transformer~\cite{Vaswani2017attention} to optimize on trajectories. 
Monotonic Quantile Network (MQN)~\cite{Bai2022monotonic} learned a distributional value function to mitigate the distribution shift problem.
Mild Offline AC (MOAC)~\cite{Huang2023mild} integrated a mild policy evaluation into an actor-critic method to bound the learned Q-values. 
Guided Offline RL (GORL)~\cite{Yang2023hundreds} proposed using expert demonstrations to guide the learning of different policy constraint intensities.
In this work, we propose a flexible framework for the model-free methods that constrain Q-functions.
ACL-QL framework supports defining different conservative levels for Q-function over each state-action pair, where Conservative Q-Learning (CQL)~\cite{Kumar2020conservative} is one special case when all conservative levels are equal.

\textbf{Model-based Offline RL.}
Recently, model-based methods~\cite{Janner2019trust,Kidambi2020morel,Yu2020mopo,Yu2021combo,Matsushima2020deployment} attract much attention for offline RL. 
They first learn the transition dynamics and reward function as a proxy environment which can be subsequently used for policy search.
%
Given the proxy environment, offline methods~\cite{ross2012agnostic,Kidambi2020morel}, or run planning and trajectory optimization like Linear Quadratic Regulator (LQR)~\cite{Tassa2012synthesis} and Monte Carlo Tree Search (MCTS)~\cite{Browne2012survey} can be directly used for controlling. 
Although model-based offline RL can be highly sample efficient, direct use of it can be challenging due to distribution shift issues. 
In this paper, we mainly focus on model-free offline RL.

\section{Preliminary}

\subsection{Deep Reinforcement Learning}
We consider the environment as a fully-observed Markov Decision Process (MDP), which is represented by a tuple $(\mS, \mA, \mP, r, \rho_0, \gamma)$. 
The MDP consists of the state space $\mS$, the actions space $\mA$, the transition probability distribution function $\mP : \mS \times \mA \times \mS \rightarrow [0,1]$, the reward function $r: \mS \times \mA \times \mS \rightarrow \bbR$, the initial state distribution $\rho_0 (s)$ and the discount factor $\gamma \in (0,1)$.
The goal is to learn a control policy $\pi(a|s)$ that maximizes the cumulative discounted return 
$g_t = \sum_{t=0}^{\infty} \gamma^{t} r(s_{t}, a_{t}, s_{t+1} | s_0 \sim \rho_{0}, a_{t} \sim \pi(\cdot | s_{t}), s_{t+1} \sim \mP(\cdot | s_{t}, a_{t}))$.
In the environment setup, although the robots are usually a time-varying system, the tasks remain static.
In the Actor-Critic framework, the learning process repeatedly alternates between the policy evaluation that computes the value function for a policy and the policy improvement that obtains a better policy from the value function.
Given a current replay buffer (dataset) $\mD = \{ (s, a, r, s') \}$ consisting of finite transition samples, 
the policy evaluation is defined as follows:
\begin{align}
\label{equ:policy_evaluation}
    \hatQ^{k+1} \leftarrow \rmarg \rmmin \limits_{Q} \bbE_{s,a,\nexts \sim \mD} 
    \left[
    \left(
    ( Q(s,a) - \mB^{\pi} \hatQ^{k}(s,a)
    \right)^{2}
    \right],
\end{align}
\dif{where $k$ is the iteration number, $Q(s,a)$ is a parametric Q-function to predict cumulative return after taking action $a$ at the state $s$, $\hatQ^{k}(s,a)$ is the frozen Q-function at iteration $k$ to calculate the target Q value, $\nexts$ is the next state,} and the Bellman operator is defined as
$\mB^{\pi} \hatQ^{k}(s,a) = r(s,a) 
    +\gamma \bbE_{\nexta \sim \pi^{k} (\nexta|\nexts)} [\hatQ^{k}(\nexts,\nexta)]$.
Note that the empirical Bellman operator $\hatB^{\pi}$ is used in practice, which backs up only one transition,
because it is difficult to contain all possible transitions $(s,a, s')$ in $\mD$, especially for continuous action space.
After approximating the Q-function, the policy improvement is performed as the following:
\begin{align}
\label{equ:policy_improvement}
    \hatP^{k+1} \leftarrow \rmarg \rmmax \limits_{\pi} \bbE_{s \sim \mD, a \sim \pi^{k}(a|s)}
    \left[
    \hatQ^{k+1}(s,a)
    \right].
\end{align}
\dif{where $\pi^{k}$ is the policy at iteration $k$ and $\hatP^{k+1}$ is the learned policy after one training step to maximize the cumulative reward.}
\subsection{Offline RL and CQL}
Compared to online RL, offline RL only allows learning from a fixed dataset $\mD$ collected by an unknown behavioral policy $\BH$, while prohibiting additional interactions with the environment.
One of the core issues in offline RL is the existence of the action distribution shift during training~\cite{Kumar2019stabilizing,Wu2019behavior,Jaques2019way,Levine2020offline}.
In the policy evaluation step with~\Cref{equ:policy_evaluation}, the Q-function is optimized using the data from $\mD$ (i.e., collected by behavioral policy $\BH$), but the target Q-values are estimated using actions sampled from current policy $\pi$.
In online RL, the interactions with the environment could provide the right feedback and thus fix such errors, yet it is impossible for offline RL.
To mitigate the distribution shift, offline RL methods are usually designed to constrain the learned policy $\pi$ to be close to the behavioral policy $\BH$ in different ways.

Conservative Q-Learning (CQL)~\cite{Kumar2020conservative}, following the pessimism principle but different from other policy-constrained methods~\cite{Jaques2019way,Wu2019behavior,Siegel2020keep}, 
proposes adding regularization terms in the policy evaluation step and learning a conservative Q-function.
More specifically, the optimization problem of CQL is the following:
\begin{align}
\label{equ:CQL}
    \rmmin \limits_{Q} & \rmmax \limits_{\mu} \alpha 
    \left(
    \bbE_{s \sim \mD, a \sim \mu(a|s)} \left[ Q(s,a) \right]
    \right.
    \notag
    \\
    & \left. - \bbE_{s \sim \mD, a \sim \hatP_{\beta}(a|s)} \left[ Q(s,a) \right]
    \right)
    \\
    & + \frac{1}{2} \bbE_{s,a,\nexts \sim \mD}
    \left[
    \left( Q(s,a) - \hatB^{\pi} \hatQ^{k}(s,a) \right)^2
    \right] + \mR(\mu), \notag
\end{align}

where $\alpha$ is a scale factor and $\hatP_{\beta}$ is the empirical behavioral policy.
The added regularization terms aim to squeeze down the Q-value of the actions from a particular distribution $\mu(a | s)$ and raise the Q-value under the empirical behavioral policy $\hatP_{\beta}$. 
\dif{This choice of distribution $\mu$ determines which distribution the learned policy $\pi$ will be far from.}
By setting $\mu = \pi$ and $\mR(\mu)=-D_{KL}(\mu, Unif(a))$, \dif{where $Unif(\cdot)$ is the uniform distribution}, CQL learns a lower bound on the state value function of the policy $\pi$ and achieves impressive results on offline RL benchmarks.

\section{Adaptive Conservative Level in Q-Learning}
\label{sec:ACL-QL_framework}

In this section, we propose a framework, Adaptive Conservative Level in Q-Learning (ACL-QL), which enables more flexible control over the conservative level of Q-function, compared to CQL~~\cite{Kumar2020conservative} that uses two separate terms to lift up the Q-values of in-dataset transitions and pull down the Q-values of OOD actions with a fixed hyperparameter $\alpha$.
The core idea behind the ACL-QL framework is that given a suitable range of
the conservative level of the Q-function, how to \emph{adaptively} lift the Q-values for good transitions up and pull down the Q-values for bad transitions in the range.
Without loss of generality, we can consider the dataset collected by the behavioral policy usually contains data with both high and low returns, even though the behavioral policy is a random policy.
At the same time, among the actions sampled from a particular distribution $\mu$, there are also (relatively) good and bad actions instead of all actions having the same returns.
In that case, we need a more flexible and fine-grained control method to constrain the Q-function for each state-action pair.
Toward our goal, we propose to use two adaptive weight functions $w_{\mu} (s,a)$ and $w_{\BH} (s,a)$ to control the conservative level over a particular distribution $\mu$ and empirical behavioral policy $\hatP_{\beta}$, respectively.
By extending the CQL framework~\cite{Kumar2020conservative}, now the family of optimization problems of our ACL-QL framework is presented below:
\begin{align}
\label{equ:ACL-QL}
    \rmmin \limits_{Q} & \rmmax \limits_{\mu} 
    \left(
    \bbE_{s \sim \mD, a \sim \mu(a|s)} \left[ w_{\mu}(s,a) \cdot Q(s,a) \right] \right.
    \notag
    \\
    & \left. - \bbE_{s \sim \mD, a \sim \hatP_{\beta}(a|s)} \left[ w_{\BH}(s,a) \cdot Q(s,a) \right] 
    \right)
    \\
    & + \frac{1}{2} \bbE_{s,a,\nexts \sim \mD}
    \left[
    \left( Q(s,a) - \hatB^{\pi} \hatQ^{k}(s,a)  \right)^2
    \right] + \mR(\mu). \notag
\end{align}
Note that the form of the adaptive weight functions $w_{\mu} (s,a)$ and $w_{\BH} (s,a)$ is not fixed and can be customized according to different situations. 
It is their arbitrary form that supports us in shaping the Q-function more finely.
%
In the following, we discuss the conditions about how we can adjust the conservative level of ACL-QL compared to the ordinary Q-function and CQL~\cite{Kumar2020conservative}, and the properties that the adaptive weight functions $w_{\mu} (s,a)$ and $w_{\BH} (s,a)$ should have.  
%
\begin{figure}
    \begin{center}
    \includegraphics[width=0.5\textwidth]{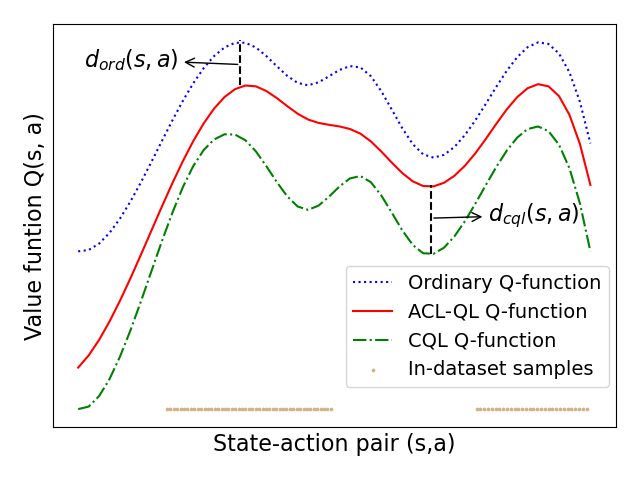}
    \end{center}
    \caption{ACL-QL can limit the Q-values into the interval between the ordinary Q-values and CQL Q-values by using $d_{ord}(s,a)$ and $d_{cql}(s,a)$ in Propositions~\ref{prop:ACL-QL_1} and~\ref{prop:ACL-QL_3} respectively.}
    \label{fig:method_example}
\end{figure}

Figure~\ref{fig:method_example} provides a schematic illustration for the following Propositions~\ref{prop:ACL-QL_1} and~\ref{prop:ACL-QL_3}, which are the conditions on how to limit the learned Q-function lies between the ordinary Q-function and the over-conservative Q-function in CQL.

First, we list the condition on which ACL-QL is more conservative than the ordinary Q-function in Proposition~\ref{prop:ACL-QL_1}.

\begin{prop}
\label{prop:ACL-QL_1}
(The conservative level of ACL-QL).
For any $\mu$ with $\rmsupp \mu \subset \rmsupp \hatP_{\beta}$, without considering the sampling error between the empirical $\hatB^{\pi} \hatQ$ and ordinary Bellman backups $\mB^{\pi} \hatQ$, the conservative level of ACL-QL can be controlled over the Q-values. 
The difference $d_{ord} (s,a)$ between the ordinary Q-function $Q^{\pi}$ and the learned Q-function $\hatQ^{\pi}$ is
\begin{align}
\label{equ:ACL-QL_theo_1_1}
    \forall s \in \mD, d_{ord} (s,a) & = Q^{\pi} - \hatQ^{\pi} \notag \\
    & =  \frac{ w_{\mu} \cdot \mu - w_{\BH} \cdot \pi_{\beta}}{\pi_{\beta}}.
\end{align}
\end{prop}
The proof is provided in Appendix I.
When $d_{ord} (s,a) > 0$, the learned Q-function $\hatQ^{\pi}$ is more conservative than the ordinary Q-function $Q^{\pi}$ point-wise.
As a special instance of our ACL-QL framework, CQL~\cite{Kumar2020conservative} proposes to constrain the conservative level over the excepted V-values. 
And CQL performs the optimization for all state-action pairs with the same weight $\alpha$, which may be too rigid and over-conservative for some scenarios. 
Now, if we set $d_{ord} (s,a) < 0$ in~\Cref{equ:ACL-QL_theo_1_1}, we can get the conditions on which ACL-QL is less conservative than the ordinary Q-function.
The Q-values of these corresponding transitions will be lifted to high values, and it is beneficial when the users believe these transitions have high qualities and want the learned policy to mimic them.

We also show that ACL-QL bounds the gap between the learned Q-values and ordinary Q-values with the consideration of the sampling error between the empirical $\hatB^{\pi} \hatQ$ and actual Bellman operator $\mB^{\pi} \hatQ$.
Following~\cite{Auer2008near,Osband2016deep,Kumar2020conservative}, the error can be bounded by leveraging the concentration properties of $\hatB^{\pi}$.
We introduce the bound in brief here:
with high probability $\geq 1 - \delta, | \hatB^{\pi} \hatQ - \mB^{\pi} \hatQ|(s,a) \leq \frac{C_{r, \mP, \delta}}{\sqrt{|\mD (s,a)|}}, \forall s, a \in \mD$, where $C_{r, \mP, \delta}$ is a constant relating to the reward function $r (s,a)$, environment dynamic $\mP (\cdot |s,a)$, and a dependency $\sqrt{log(1 / \delta)}$ where $\delta \in (0, 1)$.

\begin{prop}
\label{prop:ACL-QL_2}
(ACL-QL bounds the gap between the learned Q-values and ordinary Q-values).
Considering the sampling error between the empirical $\hatB^{\pi} \hatQ$ and ordinary Bellman backups $\mB^{\pi} \hatQ$, with probability $\geq 1 - \delta$, 
the gap between the learned Q-function $\hatQ^{\pi}$ and the ordinary Q-function $Q^{\pi}$ satisfies the following inequality:

\begin{align}
    & \forall s \in \mD, a, \notag
    \\
    & \quad \hatQ^{\pi}(s,a) - Q^{\pi}(s,a) \geq h(s,a) - err(s,a), \label{equ:ACL-QL_theo_2_1_1}
    \\
    & \quad \hatQ^{\pi}(s,a) - Q^{\pi}(s,a) \label{equ:ACL-QL_theo_2_1_2}
    \leq
    h(s,a) + err(s,a),
\end{align}
where 
\begin{align}
\label{equ:ACL-QL_theo_2_2}
h(s,a) & = - \left[ (I - \gamma P^{\pi})^{-1} \frac{ w_{\mu} \cdot \mu - w_{\BH} \cdot \hatP_{\beta}}{\hatP_{\beta}} \right] (s,a), \\
err(s,a) & = \left[ (I - \gamma P^{\pi})^{-1} \frac{C_{r, \mP, \delta} R_{max}}{(1 - \gamma \sqrt{|\mD|})} \right] (s,a) \geq 0.
\end{align}
1) Thus, if $w_{\mu} (s,a) = w_{\BH} (s,a) = \alpha, \forall s \in \mD, a$, it is the case of CQL, where the $\hatV^{\pi}$ lower-bounds the $V^{\pi}$ with a large $\alpha$ instead of a point-wise lower-bound for Q-function. 
\\
2) If $h(s,a) \geq err(s,a), \exists s \in \mD, a$, with the~\Cref{equ:ACL-QL_theo_2_1_1}, the learned Q-function $\hatQ^{\pi}$ is more optimistic than the ordinary Q-function $Q^{\pi}$ in these regions.
\\
3) If $h(s,a) \leq - err(s,a), \exists s \in \mD, a$, with the~\Cref{equ:ACL-QL_theo_2_1_2}, the learned Q-function  $\hatQ^{\pi}$ is more conservative than the ordinary Q-function $Q^{\pi}$ in these regions.
\end{prop}

Note that the term $err(s,a)$ is a positive value for any state-action pair.
Given the bounds in Proposition~\ref{prop:ACL-QL_2}, instead of only knowing the size relationship (i.e., more or less conservative), 
we can control the fine-grained range of the gap more precisely by carefully designing $w_{\mu} (s,a)$ and $w_{\BH} (s,a)$.

Besides the comparison to the ordinary Q-function, we also give a theoretical discussion about the comparison to CQL~\cite{Kumar2020conservative} in the following.

\begin{prop}
\label{prop:ACL-QL_3}
(The conservative level compared to CQL).
For any $\mu$ with $\rmsupp \mu \subset \rmsupp \hatP_{\beta}$,
given the Q-function learned from CQL is $\hatQ^{\pi}_{CQL}(s,a) = Q^{\pi} - \alpha \frac{\mu - \pi_{\beta}}{\pi_{\beta}}$, 
similar to Proposition~\ref{prop:ACL-QL_1}, the conservative level of ACL-QL compared to CQL can be controlled over the Q-values.
The difference $d_{cql}(s,a)$ between the learned Q-function $\hatQ^{\pi}$ and the CQL Q-function $\hatQ^{\pi}_{CQL}$ is
\begin{align}
\label{equ:ACL-QL_theo_3_1}
    \forall s \in \mD, d_{cql}(s,a) & = \hatQ^{\pi} - \hatQ^{\pi}_{CQL} \notag \\ 
    & = \frac{(\alpha - w_{\mu}) \mu - (\alpha - w_{\BH}) \pi_{\beta}}{\pi_{\beta}}.
\end{align}
\end{prop}
When $d_{cql} (s,a) > 0$, the learned Q-function $\hatQ^{\pi}$ is less conservative than the CQL Q-function $\hatQ^{\pi}_{CQL}$ point-wise.
The Q-values of these corresponding transitions will be higher than $\hatQ^{\pi}_{CQL}$, and it is beneficial when the users believe CQL is over-conservative on these transitions.
With the help of Propositions~\ref{prop:ACL-QL_1} and~\ref{prop:ACL-QL_3}, we can easily limit the range of the learned Q-values into a suitable interval, which is between the ordinary Q-values and CQL Q-values.

Besides the discussion about the conservative level of ACL-QL, to depict the property that a good action should have a higher Q-value than a bad action, we formally define the monotonicity of the adaptive weight functions as follows:
\begin{defi}
\label{defi:ACL-QL_1}
(The monotonicity of the adaptive weight functions). For any state $s \in \mD$, the monotonicity of the adaptive weight functions is defined as that a good action $a \in \mA$ with a higher ordinary Q value has a lower $w_{\mu}$ value and higher $w_{\BH}$ value:
\begin{align}
\label{equ:ACL-QL_defi_1}
    & \forall s_i, s_j \in \mD, a_i, a_j \in \mu(a|s), \notag \\
    & \quad w_{\mu} (s_i, a_i) - w_{\mu} (s_j, a_j)  \propto  Q^{*}(s_j, a_j) - Q^{*}(s_i, a_i). \\
    & \forall s_i, s_j \in \mD, a_i, a_j \in \hatP_{\beta}(a|s), \notag \\
    & \quad w_{\BH} (s_i, a_i) - w_{\BH} (s_j, a_j)  \propto  Q^{*}(s_i, a_i) - Q^{*}(s_j, a_j),
\end{align}
\end{defi}
where $Q^{*}$ is the optimal Q-function.
In Definition~\ref{defi:ACL-QL_1}, we use the optimal Q-function which is a natural ideal metric to measure the action quality to define the monotonicity.
However, it is an ill-posed problem since if we know the optimal Q-function, we can directly train an optimal policy and solve the problem. 
And we also do not know what the optimal proportional relationship should be.
In the next section~\ref{sec:ACL-QL}, we construct a monotonicity loss using transition quality measurements to approximate this property for ACL-QL.
All the proofs are provided in Appendix I.

\section{ACL-QL with Learnable Weight Functions}
\label{sec:ACL-QL}

In this section, derived from the theoretical discussion, we propose one practical algorithm, ACL-QL, in the guidance of theoretical conditions with a learnable neural network as adaptive weight functions. 
To adaptively control the conservative level, there are three steps for ACL-QL.
Firstly, we preprocess the fixed dataset to calculate the relative transition quality measurements.
Then, with the help of the transition quality measurements, we construct a monotonicity loss to maintain the monotonicity.
We also add surrogate losses to control the conservative level of ACL-QL.
Lastly, we train the adaptive weight network, actor network, and critic network alternatively.

\subsection{Transition Quality Measurements}

To seek a replacement for the optimal Q-function $Q^{*}$ in Definition~\ref{defi:ACL-QL_1} for offline DRL, we first preprocess the static dataset and calculate the relative transition quality measurements for each state-action pair.
Since ACL-QL should be able to be applied to different kinds of tasks with different magnitudes of rewards, it is better to use a normalized value ranging from (0, 1) to measure the data quality.

The discounted Monte Carlo return $g(s,a)$ is an unbiased way of approximating the Q-values.
However, because we do not know the coverage of the dataset (i.e., an action contained in a bad trajectory in given datasets still has the potential to achieve higher performance within other unseen trajectories), the Monte Carlo returns may have high variances, especially when the number of steps is large.

To reduce the high variances, we also consider the immediate rewards $r(s,a)$ which are 1-step Monte Carlo returns and thus have the smallest variances.
Note that only the immediate rewards $r(s,a)$ are also not accurate since our goal is to achieve higher accumulated returns, and the greedy algorithm usually leads to a sub-optimal policy for complex tasks.

Combining the above methods, for the state-action pair in the static dataset, we define relative transition quality $m(s, a)$ by combining both the whole discounted Monte Carlo returns and the single-step rewards to approximate the data quality:

\begin{align}
\label{equ:In_RTQ}
     & \forall (s, a) \in \mD, \notag \\
     & m(s, a) = \lambda \cdot g_{norm}(s, a) + (1 - \lambda) \cdot r_{norm}(s, a),
\end{align}
where $g_{norm} = \frac{g_{cur} - g_{min}}{g_{max} - g_{min}}$ and $r_{norm} = \frac{r_{cur} - r_{min}}{r_{max} - r_{min}}$.
The $g_{min}, g_{max}, r_{min}, r_{max}$ are the minimum and maximum values of the Monte Carlo returns of the whole trajectory and the single-step rewards in the dataset, respectively.
Note that the range of $m(s, a)$ is $(0, 1)$, and 
the higher $m(s, a)$ indicates that we set a pseudo-label showing $a$ is a better action.
Lines 3-4 in Algorithm~\ref{algo:pseudo-ACL-QL} show the process of calculating the relative transition quality.
$\lambda$ is a hyperparameter enabling different combinations of the Monte Carlo returns of the whole trajectory and the single-step rewards.
We build an ablation study on different $\lambda$ and N-step SARSA returns in Section~\ref{sec: ab_study}, and find that different $\lambda$ values affect the final performance, but setting it to 0.5 is a simple and good choice that works well in all environments and datasets. 
Therefore, we chose to use 0.5 in our implementation.

For the OOD actions, since they do not appear in the dataset and their single rewards and Monte Carlo returns are not available, 
we choose to use the Euclidean distance $L_2$, an intuitive and common way to approximate the uncertainties, between the OOD action $a_{\mu}$ and action $a_{in}$ in the dataset with the same state $s$.
A larger distance means that this OOD action generates a higher risk, i.e., we cannot trust its Q value too much, and hence we should suppress its Q value more.

Besides the $L_2$ distance, we also consider the quality of the corresponding in-dataset action. 
If the in-dataset action is actually a "good" one, even if the $L_2$ distance is high, we can still trust the OOD action to some extent. 
Combining the above ideas, we define relative transition quality $m(s, a_{\mu})$ for OOD state-action pairs:

\begin{align}
\label{equ:OOD_RTQ}
     & \forall (s, a_{in}) \in \mD, a_{\mu} \in \mu(a|s), \notag \\ 
     & \quad m(s, a_{\mu}) = T \left( m(s, a_{in}), L_2(a_{\mu}, a_{in}) \right),
\end{align}
where $T(x_1, x_2) = \frac{1}{2} (x_1 - \frac{1}{2} x_2 + 1)$ is a translation function to shift and scale the~\Cref{equ:OOD_RTQ} to make the range of $m(s, a_{\mu})$ is also $(0, 1)$.
The higher $m(s, a_{\mu})$ indicates that we set a pseudo-label showing $a_{\mu}$ is a better action.
Note that $a_{\mu}$ are sampled for each batch, and thus $m(s, a_{\mu})$ are calculated over every training batch.
Line 15 in Algorithm~\ref{algo:pseudo-ACL-QL} shows the process of calculating the relative transition quality.
We argue that the above~\Cref{equ:In_RTQ,equ:OOD_RTQ} is only a simple and effective way to be the replacement of the optimal Q-function, and it can serve as a baseline for future algorithms.
We believe that many other methods, including the "upper envelope" in BAIL~\cite{Chen2020bail} and uncertainty estimation in~\cite{Yu2020mopo}, also have the potential to be incorporated in ACL-QL.

\begin{algorithm}[tp]
\caption{ACL-QL}
\label{algo:pseudo-ACL-QL}
\begin{algorithmic}[1]
\State Given offline dataset $\mD$
\State Initialize behavioral policy $\BH$, policy $\pi$, Q-function $Q$ and adaptive weight functions $w_{\mu} (s,a)$ and $w_{\BH} (s,a)$
\State // Transition Quality Measurements
\State Calculate relative transition quality $m(s,a), \forall (s, a) \in \mD$ using~\Cref{equ:In_RTQ}
\State // Train behavioral policy $\BH$
\For{Each gradient step $i$}
    \State Sample transition $(s, a, s') \sim \mD$
    \State Update behavioral policy $\BH$ using~\Cref{equ:cloning_loss}
\EndFor
\State // Train policy $\pi$, Q-function $Q$ and adaptive weight functions $w_{\mu} (s,a)$ and $w_{\BH} (s,a)$
\For{Each gradient step $i$}
    \State Sample transition $(s,a, s') \sim \mD$
    \State Sample behavioral policy action $s \sim \mD, a \sim \hatP_{\beta}(a|s)$
    \State Sample current policy action $s \sim \mD, a \sim \pi(a|s)$
    \State Calculate relative transition quality $m(s, a) \ \forall s \in \mD, a \sim \hatP_{\beta}(a|s)$ or $a \sim \pi(a|s)$ using~\Cref{equ:OOD_RTQ}
    \State Update Q-function $Q$ using~\Cref{equ:ACL-QL-problem}
    \State Update $w_{\mu} (s,a)$ and $w_{\BH} (s,a)$ using~\Cref{equ:ACL-QL-problem}
    \State Update policy $\pi$ using~\Cref{equ:ACL-QL-problem}
\EndFor
\end{algorithmic}
\end{algorithm}

\subsection{Optimization to Maintain the Monotonicity}

Suppose we present the functions $w_{\mu}, w_{\BH}$ as deep neural networks, the key point is how to design the loss functions.
Given the relative transition quality $m(s,a)$ for both the in-dataset transitions and OOD actions, we firstly construct a monotonicity loss to main the monotonicity as stated in Definition~\ref{defi:ACL-QL_1}.
We use Mean Squared Error (MSE) for computation efficiency and simplicity.
The monotonicity loss is defined as the following:
\begin{align}
\label{equ:ACL-QL-mono-loss}
    & \forall (s_i, a_i), (s_j, a_j) \notin \mD, (s_k, a_k), (s_l, a_l) \in \mD, \notag
    \\ 
    & \mL_{mono}(w_{\mu}, w_{\BH}) = 
    \lVert 
    \left( \sigma(w_{\mu} (s_i, a_i)) - \sigma(w_{\mu} (s_j, a_j)) \right) \notag
    \\
    & \qquad \qquad \qquad - \left(\sigma(m(s_j, a_j)) - \sigma(m(s_i, a_i)) \right)
    \rVert_{2}^{2} \notag
    \\
    & \qquad \qquad \qquad + \lVert 
    \left( \sigma(w_{\BH} (s_k, a_k)) - \sigma(w_{\BH} (s_l, a_l)) \right) \notag
    \\
    & \qquad \qquad \qquad - \left(\sigma(m(s_k, a_k)) - \sigma(m(s_l, a_l)) \right)
    \rVert_{2}^{2},
\end{align}
where we use the softmax operation $\sigma(\cdot)$ over the current batch of the training data to unify the orders of magnitude between the adaptive weights and transition quality measurements. 

\subsection{Optimization to Control the Conservative Level}

Recapping the~\Cref{equ:ACL-QL_theo_1_1,equ:ACL-QL_theo_3_1}, we have known the conditions for different conservative levels, and thus, we can adapt them to the loss functions.
The goal is to learn a mild Q-function ranging in the interval between the ordinary Q-function and the CQL Q-function with a large $\alpha$. 
More specifically, by multiplying the denominators in~\Cref{equ:ACL-QL_theo_1_1,equ:ACL-QL_theo_3_1} simultaneously, we incorporate the conditions into the hinge losses as shown in the following:
\begin{align}
    \mL_{ord}(w_{\mu}, w_{\BH}) & = max(0, w_{\BH} \cdot \BH \notag 
    \\
    & - w_{\mu} \cdot \mu + d_{ord} \cdot \BH), \label{equ:ACL-QL-loss1} 
    \\
    \mL_{cql}(w_{\mu}, w_{\BH}) & = max(0, (w_{\mu} - \alpha) \cdot \mu \notag
    \\
    & - (w_{\BH} - \alpha) \cdot \BH + d_{cql} \cdot \BH), \label{equ:ACL-QL-loss2}
\end{align}
where $d_{ord}$ and $d_{cql}$ are used to control the differences of the conservative level compared to the ordinary Q-function and CQL, respectively.
A larger $d_{ord}$ means a more conservative Q-function than the ordinary Q-function, while a larger $d_{cql}$ represents a less conservative Q-function than CQL.
However, if we set $d_{ord}$ and $d_{cql}$ as hyperparameters, they are difficult to tune and not adaptive enough for different tasks and datasets.
%
%

Considering the monotonicity property in Definition~\ref{defi:ACL-QL_1} that a good action with a larger $m(s,a)$ should have less conservative (larger) Q-value, which is the equivalent of a smaller $d_{ord}$ and a larger $d_{cql}$,
we leverage the relative transition quality $m(s,a)$ to calculate $d_{ord}$ and $d_{cql}$ automatically.
Recall that the range of relative transition quality $m(s,a)$ is $(0,1)$, while the $d_{ord}$ and $d_{cql}$ are the Q-value gaps compared to the ordinary Q-values and CQL Q-values.
For different tasks and datasets, the overall magnitudes of Q-values are not the same. 
Hence, we have to shift and scale the range of $m(s,a)$ to more suitable intervals for each dataset.

In practice, we try to scale $m(s,a)$ by (1) the mean value $r_{mean}$ of all single-step rewards over the whole dataset, (2) the maximum single-step reward $r_{max}$ over the whole dataset and (3) the upper bound of Q-values $Q_{up} = \frac{r_{max}}{1 - \gamma}$.
We find that $r_{mean}$ are so small that it is difficult to distinguish significantly between good and bad transitions, especially for random datasets where the data qualities are not good.
On the other hand, the $Q_{up}$ is too large for the Q-value gaps $d_{ord}$ and $d_{cql}$.
Imagine that the difference between two Q-values is the upper bound of all Q-values, 
it would relax the conditions too much in~\Cref{equ:ACL-QL_theo_1_1,equ:ACL-QL_theo_3_1}, and lead to the invalidation of the restriction.
The choice $r_{max}$ works better than the other two choices and the calculation of the $d_{ord}$ and $d_{cql}$ is as the following:

\begin{align}
    d_{ord} (s, a) & = (1 - m(s, a)) \cdot r_{max}, \label{equ:d_ord} \\
    d_{cql} (s, a) & = m(s, a) \cdot r_{max}. \label{equ:d_cql}
\end{align}

Nevertheless, during the optimization for $\mL_{ord}$ and $\mL_{cql}$, we find it is prone to cause arithmetic underflow since the $log_{\mu}, log_{\BH}$ is usually very small like -1000, then the resulting $\mu(a|s)$ and $\BH(a|s)$ become 0 after exponentiation operation.
To avoid the arithmetic underflow problem, we use a necessity of~\Cref{equ:ACL-QL-loss1,equ:ACL-QL-loss2} to form surrogate losses.
\begin{lemma}
\label{lemma:ACL-QL}
For $x > 0$, $\ln{x} \leq x - 1$. $\ln{x} = x - 1$ if and only if $x = 1$.
\end{lemma}
The proof is provided in Appendix I.
Then, the resulting surrogate losses are the following:
\begin{align}
    & \mL_{ord}(w_{\mu}, w_{\BH}) = max(0, w_{\BH} \cdot (\ln{\BH} + 1) \notag
    \\
    & \quad \quad - w_{\mu} \cdot (\ln{\mu} + 1) + d_{ord} \cdot (\ln{\BH} + 1)), \label{equ:ACL-QL-sur-loss1} 
    \\
    & \mL_{cql}(w_{\mu}, w_{\BH}) = max(0, (w_{\mu} - \alpha) (\ln{\mu} + 1) \notag
    \\
    & \quad \quad - (w_{\BH} - \alpha) (\ln{\BH} + 1) + d_{cql} (\ln{\BH} + 1)). \label{equ:ACL-QL-sur-loss2}
\end{align}

As shown in~\Cref{equ:ACL-QL-loss1,equ:ACL-QL-loss2}, we need to calculate the log probability $\ln{\BH}$ from the behavioral policy. 
Thus, we use a neural network with a Gaussian head to represent the behavioral policy $\BH$ and train it using behavioral cloning as follows:
\begin{align}
\label{equ:cloning_loss}
    \mL_{\beta}= \mathbb{E}_{(s, a) \sim \mD}[(\BH(s) - a)^2].
\end{align}
The training process corresponds to lines 5-9 in Algorithm~\ref{algo:pseudo-ACL-QL}.

\subsection{Final Object}

To learn a conservative Q-function, the adaptive weight should be positive, and thus, we add a regularization term as follows:
\begin{align}
    \mL_{pos}(w_{\mu}, w_{\BH}) & = max(0, - w_{\mu}) + max(0,  - w_{\BH}). \label{equ:ACL-QL-pos_loss}
\end{align}
The final object for training the adaptive weight functions $w_{\mu}, w_{\BH}$ is:
\begin{align}
    \mL(w_{\mu}, w_{\BH}) & = \mL_{ord}(w_{\mu}, w_{\BH}) + \mL_{cql}(w_{\mu}, w_{\BH}) \notag
    \\
    & + \mL_{mono}(w_{\mu}, w_{\BH}) + \mL_{pos}(w_{\mu}, w_{\BH}).  \label{equ:ACL-QL-final_loss}
\end{align}

ACL-QL is built on the top of the CQL~\cite{Kumar2020conservative}, which sets $\mu = \pi$ and $\mR(\mu)=-D_{KL}(\mu, Unif(a))$ in~\Cref{equ:ACL-QL} in Section~\ref{sec:ACL-QL}.
The optimization problem becomes:
\begin{align}
\label{equ:ACL-QL-problem}
    \rmmin \limits_{Q} \rmmax \limits_{\pi} \rmmin \limits_{w_{\mu}, w_{\BH}}
    \left(
    \bbE_{s \sim \mD, a \sim \pi(a|s)} \left[ w_{\mu}(s,a) \cdot Q(s,a) \right] \right. \notag
    \\
    \left. - \bbE_{s \sim \mD, a \sim \hatP_{\beta}(a|s)} \left[ w_{\BH}(s,a) \cdot Q(s,a) \right]
    \right) \notag
    \\
    + \frac{1}{2} \bbE_{s,a,\nexts \sim \mD}
    \left[
    \left( Q(s,a) - \hatB^{\pi} \hatQ^{k}(s,a) \right)^2
    \right] \notag
    \\
    - \mD_{KL} (\pi, Unif(a)) + \mL(w_{\mu}, w_{\BH}).
\end{align}
 
During the training process, which is shown in lines 10-19 in Algorithm~\ref{algo:pseudo-ACL-QL}, we add weight neural networks $w_{\mu}, w_{\BH}$ and train the weight networks, Q networks, and policy networks alternatively.

\begin{table}[t]
\caption{Implementation Details of ACL-QL.}
\label{tab:para_ACL-QL}
\centering
\resizebox{0.98\columnwidth}{!}{
    \begin{tabular}{cll}
    \toprule
     & Hyperparameter & Value \\ 
    \midrule
    \multirow{13}*{\shortstack{ACL-QL\\Hyperparameter}} 
    & Critic learning rate & 3e-4 \\ 
    & Actor learning rate & 1e-5 \\ 
    & Target update rate & 5e-3 \\ 
    & Optimizer & Adam~\cite{Kipf2016semi} \\ 
    & Batch size & 256 \\ 
    & Discount factor & 0.99 \\ 
    & Lagrange for $\alpha$ & False \\ 
    & Training steps & 1.1 M \\
    & Evaluation frequency & 1000 \\
    & Evaluation episodes & 10 \\
    & Weight function learning rate & 3e-4 \\ 
    & Behavioral policy training steps & 0.1 M \\ 
    \midrule
    \multirow{11}*{\shortstack{ACL-QL \\ Network\\Architectures}} 
    & Actor type & Tanh \& Gaussian \\ 
    & Actor hidden layers & 3 \\ 
    & Actor hidden dim & [256, 256, 256] \\ 
    & Actor activation function & ReLU~\cite{Agarap2018deep} \\
    & Critic hidden layers & 3 \\ 
    & Critic hidden dim & [256, 256, 256] \\
    & Critic activation function & ReLU \\ 
    & Weight function hidden layers & 3 \\ 
    & Weight function hidden dim & [256, 256, 256] \\
    & Weight function output dim & 2 \\ 
    & Weight function activation function & ReLU \\ 
    \bottomrule
    \end{tabular}
}
\end{table}

\begin{table*}[t]\scriptsize
    \caption{Normalized average accumulated rewards on 12 D4RL Gym-MuJoCo-v2 environments. We reported the results of all baselines from the D4RL whitepaper and three original papers. We reported the mean and standard deviation values of ACL-QL over 6 random seeds. The best results are bold, and the second-best results are underlined.}
    \label{tab:main_gymv2_results}
    \centering
    \resizebox{\textwidth}{!}{
        \begin{NiceTabular}{l|cccccccc|c}
        \toprule
        {\bf Task Name} & {\bf AWAC} & {\bf OnestepRL} & {\bf TD3+BC} & {\bf IQL} & {\bf CQL} & {\bf \dif{GORL}} & {\bf \dif{MQN-CQR}} & {\bf \dif{MOAC}} & {\bf ACL-QL} \\
        \midrule
        halfcheetah-m-e-v2 & 36.8 & \textbf{93.4} & 90.7 & 86.7 & \underline{91.6} & 88.2 & 71.1 & 87.2 & 87.4 $\pm$ 7.5 \\
        halfcheetah-m-r-v2 & 40.5 & 38.1 & 44.6 & 44.2 & 45.5 & 46.5 & 45.3 & \underline{50.3} & \textbf{55.9} $\pm$ 6.7 \\
        halfcheetah-m-v2 & 37.4 & 48.4 & 48.3 & 47.4 & 44.0 & 51.7 & 45.1 & \underline{54.3} & \textbf{69.8} $\pm$ 2.5 \\
        halfcheetah-r-v2 & 2.2 & 6.9 & 11.0 & - & 18.6 & 16.2 & \textbf{32.6} & 24.0 & \underline{28.5} $\pm$ 4.6 \\
        \midrule
        halfcheetah-sum & 116.9 & 186.8 & 194.6 & 178.3 & 199.7 & 202.6 & 194.1 & \underline{215.8} & \textbf{241.6} \\
        \midrule
        hopper-m-e-v2 & 80.9 & 103.3 & 98.0 & 91.5 & 105.4 & 88.8 & \textbf{113.0} & 102.1 & \underline{107.2} $\pm$ 4.9 \\
        hopper-m-r-v2 & 37.2 & 97.5 & 60.9 & 94.7 & 95.0 & 74.7 & 95.6 & \underline{98.0} & \textbf{99.3} $\pm$ 3.9 \\
        hopper-m-v2 & 72.0 & 59.6 & 59.3 & 66.3 & 58.5 & 64.2 & \underline{94.7} & 83.4 & \textbf{97.9} $\pm$ 6.4 \\
        hopper-r-v2 & 9.6 & 7.8 & 8.5 & - & 9.3 & 15.8 & 13.2 & \underline{30.3} & \textbf{33.5} $\pm$ 1.2 \\
        \midrule
        hopper-sum & 199.7 & 268.2 & 226.7 & 252.5 & 268.2 & 243.5 & \underline{316.5} & 313.8 & \textbf{337.9} \\
        \midrule
        walker-m-e-v2 & 42.7 & \underline{113.0} & 110.1 & 109.6 & 108.8 & 109.6 & 112.1 & 111.4 & \textbf{113.4} $\pm$ 6.3 \\
        walker-m-r-v2 & 27.0 & 49.5 & 81.8 & 73.9 & 77.2 & 79.3 & 52.3 & \underline{90.0} & \textbf{96.5} $\pm$ 4.2 \\
        walker-m-v2 & 30.1 & 81.8 & \underline{83.7} & 78.3 & 72.5 & 83.5 & 80.0 & \textbf{86.7} & 79.3 $\pm$ 8.2 \\
        walker-r-v2 & 5.1 & 6.1 & 1.6 & - & 2.5 & 3.6 & \textbf{22.6} & 4.4 & \underline{22.3} $\pm$ 1.2 \\
        \midrule
        walker-sum & 104.9 & 250.4 & 277.2 & 261.8 & 261.0 & 276.0 & 267.0 & \underline{292.5} & \textbf{311.5} \\
        \bottomrule
        \end{NiceTabular}
    }
\end{table*}

\begin{table*}[t]\scriptsize
    \caption{The sum of the normalized average accumulated rewards on all 3 different datasets for each task on D4RL Adroit environments. The results of ACL-QL are over 6 random seeds.}
    \label{tab:main_adroit_results}
    \centering
    \resizebox{\textwidth}{!}{
        \begin{NiceTabular}{l|ccccccccc|c}
        \toprule
        {\bf Task Name} & {\bf BC} & {\bf SAC-Off} & {\bf BEAR} & {\bf BRAC-p} & {\bf BRAC-v} & {\bf AWR} & {\bf BCQ} & {\bf aDICE} & {\bf CQL} & {\bf ACL-QL}\\
        \midrule
        door-v0-sum & 35.3 & 11.4 & 103.0 & -0.7 & -0.7 & 103.3 & 99.9 & 0.0 & \underline{111.8} & \textbf{112.2} \\
        hammer-v0-sum & \textbf{127.9} & 25.9 & \textbf{127.9} & 0.9 & 0.8 & 40.6 & 108.1 & 0.9 & 93.2 & \underline{124.1} \\
        pen-v0-sum & 176.4 & 35.9 & 131.4 & 6.2 & -4.9 & 151.3 & \underline{227.8} & -9.7 & 183.7 & \textbf{228.6} \\
        \bottomrule
        \end{NiceTabular}
    }
\end{table*}

\begin{table*}[t]\scriptsize
    \caption{Normalized average accumulated rewards on 3 D4RL Franka Kitchen environments. We reported the mean and standard deviation values of ACL-QL over 6 random seeds.}
    \label{tab:main_kitchen_results}
    \centering
    \resizebox{\textwidth}{!}{
        \begin{NiceTabular}{l|ccccccccc|c}
        \toprule
        {\bf Task Name} & {\bf BC} & {\bf SAC-Off} & {\bf BEAR} & {\bf BRAC-p} & {\bf BRAC-v} & {\bf AWR} & {\bf BCQ} & {\bf aDICE} & {\bf CQL} & {\bf ACL-QL}\\
        \midrule
        kitchen-complete-v0 & 33.8 & 15.0 & 0.0 & 0.0 & 0.0 & 0.0 & 8.1 & 0.0 & 43.8 & \textbf{51.1} $\pm$ 6.3 \\
        kitchen-mixed-v0 & 47.5 & 2.5 & 47.2 & 0.0 & 0.0 & 10.6 & 8.1 & 2.5 & 51.0 & \textbf{53.8} $\pm$ 5.1 \\
        kitchen-partial-v0 & 33.8 & 0.0 & 13.1 & 0.0 & 0.0 & 15.4 & 18.9 & 0.0 & 49.8 & \textbf{52.4 $\pm$ 5.7} \\
        \bottomrule
        \end{NiceTabular}
    }
\end{table*}

\section{Experiments}

\subsection{Implementation Details}

In ACL-QL, we represent two adaptive weight functions $w_{\mu}(s,a)$ and $w_{\BH}(s,a)$ by one neural network, which has the same network architecture as the Q-function but the output dimension is 2.
During each gradient descent step, we train the adaptive weight network, the Q networks, and the policy networks in turn.
We list all the hyperparameters and network architectures of ACL-QL in Table~\ref{tab:para_ACL-QL}.

\subsection{Comparisons to Offline RL Baselines}

We conducted all the experiments on the commonly-used offline RL benchmark D4RL~\cite{Fu2020d4rl}, which includes many task domains~\cite{Todorov2012mujoco,Brockman2016openai,Rajeswaran2017learning}
and a variety of dataset types.
The evaluation criterion is the normalized average accumulated rewards ranging from 0 to 100, where 0 represents the average returns of a policy taking random action at each step, and 100 represents the average returns of a domain-specific expert.

Aiming to provide a comprehensive comparison, we compared ACL-QL to many state-of-the-art model-free algorithms including Behavioral Cloning (BC)~\cite{Pomerleau1988alvinn}, Behavioral Cloning with best 10 \% trajectories (10 \% BC)~\cite{Pomerleau1988alvinn}, offline version of Soft Actor-Critic (SAC-Off)~\cite{Haarnoja2018soft}, Bootstrapping Error Accumulation Reduction (BEAR)~\cite{Kumar2019stabilizing}, Behavior Regularized Actor-Critic (BRAC)~\cite{Wu2019behavior}, Advantage Weighted Regression (AWR)~\cite{Peng2019advantage}, Batch-Constrained
deep Q-learning (BCQ)~\cite{Fujimoto2019off}, Algorithm for policy gradient from arbitrary experience via DICE (aDICE)~\cite{Nachum2019algaedice}, Decision Transformer (DT)~\cite{Chen2021decision},  Advantage Weighted Actor-Critic 
 (AWAC)~\cite{Nair2020accelerating}, Onestep RL~\cite{Brandfonbrener2021offline}, TD3+BC~\cite{Fujimoto2021minimalist}, Implicit Q-Learning (IQL)~\cite{Kostrikov2021implicit}, Conservative Q-Learning (CQL)~\cite{Kumar2020conservative}, \dif{Guided Offline RL (GORL)~\cite{Yang2023hundreds}, Monotonic Quantile
network with Conservative Quantile Regression (MQN-CQR)~\cite{Bai2022monotonic} and Mild Offline AC (MOAC)~\cite{Huang2023mild}.}
For the sake of fair comparisons, we directly reported the results of all baselines from the D4RL whitepaper~\cite{Fu2020d4rl} and their original papers.
To be consistent with previous works, we trained ACL-QL for 1.0 M gradient steps and evaluated for 10 episodes every 1000 training iterations.
The results are the average accumulated reward of 10 episodes over 6 random seeds~\cite{Islam2017reproducibility} and are obtained from the workflow proposed by~\cite{Kumar2021workflow}. 
More specifically, we use the offline policy evaluation method in~\cite{Kumar2021workflow}, which calculates the average Q-value of all states in the dataset and the corresponding predicted actions to choose the model checkpoint.
We select the model with the highest average Q-value as the final model.
The basic idea here is that the goal of the learned policy is to maximize the accumulative discounted return over all states, which include the states in the dataset.
Once the policy can achieve high Q-values over all states in the dataset, the current policy is a good one to a certain extent.
Note that to reflect better the fact that ACL-QL can adaptively learn good strategies for various datasets, we use the \emph{same set} of hyperparameters for all tasks on the same environment, as opposed to other baseline algorithms that adjust hyperparameters individually for each task to obtain optimal performances.
This greatly reduces the extent to which the prior knowledge of the dataset is required and reduces the difficulty and cost for deployment.

\textbf{Gym-MuJoCo Tasks.} 
We evaluate ACL-QL over the Gym-MuJoCo tasks, including "halfcheetah", "hopper" and "walker" with 4 kinds of dataset types, ranging from medium-expert mixed data to random data.
For brevity, we marked "-medium-expert", "-medium-replay", "-medium" and "random" as "-m-e", "-m-r", "-m" and "-r" respectively.
We use $\alpha=10$ for "halfcheetah" and "walker" environments and $\alpha=20$ for "hopper" environments.
Table~\ref{tab:main_gymv2_results} shows the normalized average accumulated rewards of 12 Gym-MuJoCo version-2 tasks over 6 random seeds.
We can observe that ACL-QL consistently outperforms other baselines for the sum results on HalfCheetah, Hopper, and Walker environments by a large margin,
like achieving normalized returns 337.9 for Hopper-sum compared to 268.2 from CQL~\cite{Kumar2020conservative}.
From the perspective of the dataset types, ACL-QL is a very balanced algorithm that achieves excelling results on all kinds of datasets with expert, medium, and random data.
For instance, with the same hyperparameters, on Hopper-medium-expert, ACL-QL achieves a comparable high performance of 107.2, while also gaining the best result of 33.5 on Hopper-random.
\dif{In contrast, though MQN-CQR delivers relatively higher returns on datasets with better quality, such as 113.0 on Hopper-medium-expert, it can only have lower returns on random datasets like 13.2 on Hopper-random.}

\textbf{Adroit Tasks.}
The adroit tasks~\cite{Rajeswaran2017learning} are high-dimensional robotic manipulation tasks with sparse reward, including "door", "hammer", and "pen".
Each task includes expert data and human demonstrations from narrow distributions and contains 3 types, i.e., "-cloned", "-expert", and "-human".
We use $\alpha=1$ for "pen", $\alpha=10$ for "hammer", and $\alpha=20$ for "door" environments.
Table~\ref{tab:main_adroit_results} shows the sum of the normalized average accumulated rewards on all 3 different datasets for each task over 6 random seeds.
ACL-QL delivers higher performance than other baselines on "door" and "pen" environments and achieves comparable results on "hammer" environments.
Another significant observation is that BC can gain very high results on all tasks, e.g., 127.9 on "hammer", since most data are from expert and human demonstrations following narrow distributions.

\textbf{Franka Kitchen Tasks.} 
The Franka Kitchen tasks~\cite{Gupta2019relay} include complex trajectories as the offline datasets and aim to evaluate the "stitching" ability of the agent in a realistic kitchen environment.
We use $\alpha=20$ for all kitchen tasks.
Table~\ref{tab:main_kitchen_results} shows the normalized average accumulated rewards of all 3 Kitchen tasks over 6 random seeds.
The results of BC show that all three datasets contain a portion of high-quality data, yet many baselines except CQL~\cite{Kumar2020conservative} failed directly.
We can see that ACL-QL consistently exceeds CQL, BC, and other baselines on all three kinds of datasets since ACL-QL can control the conservative level adaptively.

\begin{table*}[t]\scriptsize
    \caption{Comparisons of CQL with different conservative levels ($\alpha=1,5,10,20$) on D4RL Gym-MuJoCo-v2 environments in terms of the sum of the normalized average accumulated rewards on the medium-expert, medium-replay, medium, and random datasets over 6 random seeds.}
    \label{tab:compare_cql_sum_v2}
    \centering
    \resizebox{0.75\textwidth}{!}{
        \begin{NiceTabular}{l|cccc|c}
        \toprule
        {\bf Task Name} & {\bf CQL-1} & {\bf CQL-5} & {\bf CQL-10} & {\bf CQL-20} & {\bf ACL-QL} \\
        \midrule
        halfcheetah-v2-sum & 154.0 & \underline{198.2} & 183.1 & 181.5 & \textbf{241.6} \\
        hopper-v2-sum & 251.0 & 267.3 & \underline{277.3} & 232.4 & \textbf{337.9} \\
        walker-v2-sum & 191.2 & 278.1 & \underline{278.5} & 258.9 & \textbf{311.5} \\
        \bottomrule
        \end{NiceTabular}
    }
\end{table*}

\begin{table*}[t]\scriptsize
    \caption{Comparisons of CQL with different conservative levels ($\alpha=1,5,10,20$) on D4RL Gym-MuJoCo HalfCheetah-v2 environments in terms of normalized average accumulated rewards over 6 random seeds.}
    \label{tab:compare_cql_half_v2}
    \centering
    \resizebox{0.8\textwidth}{!}{
        \begin{NiceTabular}{l|cccc|c}
        \toprule
        {\bf Task Name} & {\bf CQL-1} & {\bf CQL-5} & {\bf CQL-10} & {\bf CQL-20} & {\bf ACL-QL} \\
        \midrule
        halfcheetah-m-e-v2 & 54.9 $\pm$ 3.7 & 78.2 $\pm$ 3.3 & 67.5 $\pm$ 4.1 & \underline{85.3} $\pm$ 3.9 & \textbf{87.4} $\pm$ 7.5 \\
        halfcheetah-m-r-v2 & 14.1 $\pm$ 1.2 & \underline{46.4} $\pm$ 2.3 & 46.7 $\pm$ 3.2 & 43.6 $\pm$ 2.8 & \textbf{55.9} $\pm$ 6.7 \\
        halfcheetah-m-v2 & \underline{56.2} $\pm$ 4.1 & 47.5 $\pm$ 3.3 & 47.1 $\pm$ 3.6 & 45.8 $\pm$ 3.3 & \textbf{69.8} $\pm$ 2.5 \\
        halfcheetah-r-v2 & \textbf{28.8} $\pm$ 2.6 & 26.1 $\pm$ 3.1 & 21.8 $\pm$ 2.7 & 6.8 $\pm$ 1.3 & \underline{28.5} $\pm$ 4.6 \\
        \midrule
        halfcheetah-sum & 154.0 & \underline{198.2} & 183.1 & 181.5 & \textbf{241.6} \\
        \bottomrule
        \end{NiceTabular}
    }
\end{table*}

\begin{figure*}[t]
    \begin{minipage}[t]{0.33\linewidth}
    \includegraphics[width=1\linewidth]{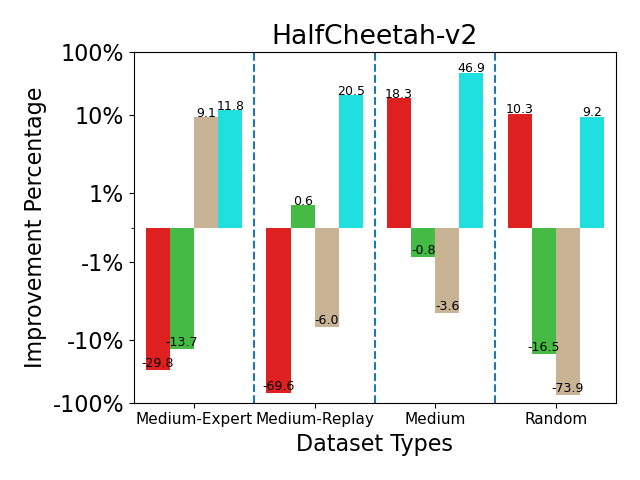}
    \end{minipage}%
    \begin{minipage}[t]{0.33\linewidth}
    \includegraphics[width=1\linewidth]{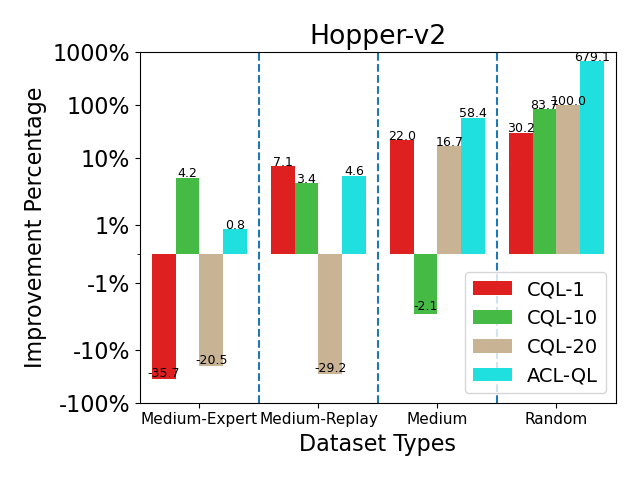}
    \end{minipage}%
    \begin{minipage}[t]{0.33\linewidth}
    \includegraphics[width=1\linewidth]{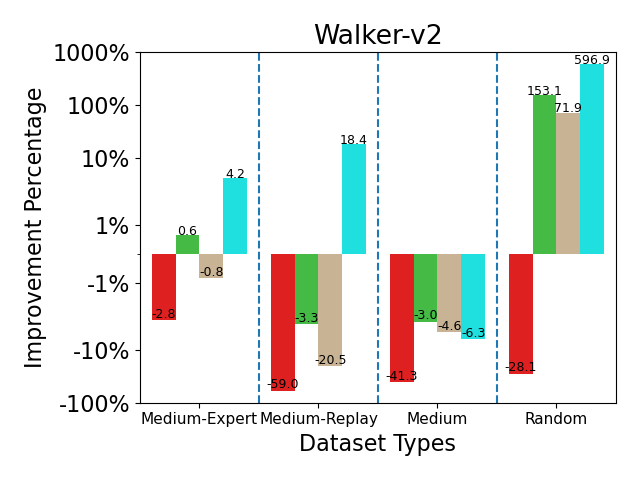}
    \end{minipage}%
\caption{To eliminate the impact of gaps in results due to the quality of the dataset itself and fairly show the comparison for different datasets, we use CQL-5 as a baseline and report the improvement percentages of the normalized results of ACL-QL and CQL with other conservative levels ($\alpha=1,10,20$) on D4RL Gym-MuJoCo-v2 environments. Note that we scale the y-axis using \textit{log} due to the huge improvements of ACL-QL on random datasets (e.g., ACL-QL achieves 33.5 on Hopper-random-v2, which is 679.1\% times higher than the results of 4.9 for CQL-5).}
\label{fig:compare_cql_all_v2}
\end{figure*}

\begin{table}[t]\scriptsize
    \caption{Comparisons of average Q-values over the datasets on HalfCheetah-v2 environments. The Q-values of SAC~\cite{Haarnoja2018soft} are from the ordinary Q-functions. We also provide the Q-values of online SAC. The Q-values of ACL-QL are between SAC and CQL-10 and are closer to the Q-values of online SAC, showing that the losses $\mL_{ord}$, $\mL_{cql}$ in~\Cref{equ:ACL-QL-loss1,equ:ACL-QL-loss2} take effect.}
    \label{tab:half-q-comparisons}
    \centering
    \resizebox{0.45\textwidth}{!}{
        \begin{NiceTabular}{l|ccc|c}
        \toprule
        {\bf Task Name} & {\bf \dif{Online SAC}} & {\bf SAC} & {\bf CQL-10} & {\bf ACL-QL} \\
        \midrule
        medium-expert & 1022.3 & 3681.4 & 676.7 & 884.3 \\
        medium-replay & 346.5 & 578.6 & 165.8 & 267.1 \\
        medium & 826.3 & 1568.6 & 273.1 & 693.8 \\
        random &  234.1  & 267.2 & -19.1 & 212.9 \\
        \bottomrule
        \end{NiceTabular}
    }
\end{table}

\begin{table*}[t]\scriptsize
    \caption{Ablation Study with the proposed losses on D4RL Gym-MuJoCo HalfCheetah-v2 environments in terms of normalized average accumulated rewards over 6 random seeds.}
    \label{tab:abl_half_results}
    \centering
    \resizebox{1.0\textwidth}{!}{
        \begin{NiceTabular}{l|cccc|cccc}
        \toprule
        & $\maL_{ord}$ & $\maL_{cql}$ & $\maL_{mono}$ & $\maL_{pos}$ & {\bf halfcheetah-m-e} & {\bf halfcheetah-m-r} & {\bf halfcheetah-m} & {\bf halfcheetah-r} \\
        \midrule
        1 & \ding{51} &  &  &  & 38.2 $\pm$ 6.4 & 26.6 $\pm$ 7.8 & 35.1 $\pm$ 3.3 & 12.3 $\pm$ 3.7 \\
        2 &  & \ding{51} &  &  & 40.3 $\pm$ 7.7 & 48.1 $\pm$ 6.1 & 47.6 $\pm$ 3.8 & 7.1 $\pm$ 2.7 \\
        3 & \ding{51} & \ding{51} &  &   & 42.6 $\pm$ 7.1 & 44.9 $\pm$ 6.5 & 61.5 $\pm$ 4.4 & 8.8 $\pm$ 3.1 \\
        4 &  &  & \ding{51} &   & 41.6 $\pm$ 8.5 & 21.5 $\pm$ 10.1 & 61.3 $\pm$ 3.9 & 6.9 $\pm$ 3.4 \\
        5 & \ding{51} & \ding{51} & \ding{51} &  & \underline{84.2} $\pm$ 6.7 & \underline{49.3} $\pm$ 6.9 & \underline{66.4} $\pm$ 4.1 & \underline{18.3} $\pm$ 4.5 \\
        \midrule
        6 & \ding{51} & \ding{51} & \ding{51} & \ding{51} & \textbf{87.4} $\pm$ 7.5  & \textbf{55.9} $\pm$ 6.7 & \textbf{69.8} $\pm$ 2.5 & \textbf{28.5} $\pm$ 4.6 \\
        \bottomrule
        \end{NiceTabular}
    }
\end{table*}

\begin{table*}[t]\scriptsize
    \caption{Ablation Study with different transition quality measurements on D4RL Gym-MuJoCo HalfCheetah-v2 environments in terms of normalized average accumulated rewards over 6 random seeds.}
    \label{tab:abl_qua_mea_half_v2}
    \centering
    \resizebox{1.0\textwidth}{!}{
        \begin{NiceTabular}{l|cccccc|c}
        \toprule
        {\bf Task Name} & {\bf $\lambda=0.0$} & {\bf $\lambda=0.3$} & {\bf $\lambda=0.7$}  & {\bf $\lambda=1.0$} & {5-step SARSA} & {10-step SARSA} & {\bf ACL-QL($\lambda=0.5$)} \\
        \midrule
        halfcheetah-m-e-v2 & 79.8 $\pm$ 6.7 & 82.3 $\pm$ 8.3 & \textbf{93.7} $\pm$ 6.9 & \underline{93.1} $\pm$ 6.5 & 83.4 $\pm$ 7.1 & 86.6 $\pm$ 7.6 & 87.4 $\pm$ 7.5 \\
        halfcheetah-m-r-v2 & 48.9 $\pm$ 6.8 & \underline{51.4} $\pm$ 4.6 & 47.8 $\pm$ 5.6 & 47.0 $\pm$ 5.3 & 47.1 $\pm$ 4.0 & 48.7 $\pm$ 4.9 & \textbf{55.9} $\pm$ 6.7 \\
        halfcheetah-m-v2 & 57.5 $\pm$ 3.7 & 63.8 $\pm$ 2.9 & 67.6 $\pm$ 3.0 & \underline{69.4} $\pm$ 3.3 & 67.1 $\pm$ 2.5 & 65.7 $\pm$ 3.1 & \textbf{69.8} $\pm$ 2.5 \\
        halfcheetah-r-v2 & \textbf{32.7} $\pm$ 4.4 & \underline{29.1} $\pm$ 4.5 & 24.2 $\pm$ 5.1 & 18.4 $\pm$ 4.9 & 25.3 $\pm$ 3.2 & 21.7 $\pm$ 3.7 & 28.5 $\pm$ 4.6 \\
        \midrule
        halfcheetah-sum & 218.9 & 226.6 & \underline{233.3} & 227.9 & 222.9 & 219.7 & \textbf{241.6} \\
        \bottomrule
        \end{NiceTabular}
    }
\end{table*}

\begin{table}[t]\scriptsize
    \caption{Comparisons of different discount factor $\gamma$ over the datasets on HalfCheetah-v2 environments in terms of normalized average accumulated rewards over 6 random seeds. }
    \label{tab:abl-gamma-comparisons}
    \centering
    \resizebox{0.45\textwidth}{!}{
        \begin{NiceTabular}{l|c|c}
        \toprule
        {\bf Task Name} & {\bf $\gamma=1.0$} & {\bf $\gamma=0.99$} \\
        \midrule
        medium-expert & \textbf{87.7} $\pm$ 6.8 & 87.4 $\pm$ 7.5 \\
        medium-replay & 45.3 $\pm$ 6.9 & \textbf{55.9} $\pm$ 6.7 \\
        medium & 65.4 $\pm$ 3.2 & \textbf{69.8} $\pm$ 2.5 \\
        random & 24.6 $\pm$ 3.7 & \textbf{28.5} $\pm$ 4.6 \\
        \bottomrule
        \end{NiceTabular}
    }
\end{table}

\subsection{Comparisons among Different Conservative Levels}

To demonstrate more intuitively that ACL-QL can adaptively control the conservative level, we implement CQL~\cite{Kumar2020conservative} locally with different $\alpha$ values ranging from ${1, 5, 10, 20}$ and compare ACL-QL to them.
The different $\alpha$ values represent different conservative levels, and a higher $\alpha$ means more conservative on the Q-function.

Table~\ref{tab:compare_cql_sum_v2} shows the sum of the normalized average accumulated rewards on "HalfCheetah", "Hopper" and "Walker" environments.
We can observe that ACL-QL can easily outperform CQL with fixed conservative levels $\alpha$ in terms of total results for all different MoJoCo-v2 environments. 
One major reason is that CQL can not adaptively adjust its conservative level according to the qualities of the datasets, leading to high performance on one dataset but dropping dramatically on the others. 

In Figure~\ref{fig:compare_cql_all_v2}, to eliminate the impact of gaps in results due to the quality of the dataset itself and fairly show the comparison for different datasets, we use CQL-5 as a baseline and plot the improvement percentages of the normalized results of ACL-QL and CQL with other conservative levels $\alpha=1,10,20$ on all Gym-MuJoCo-v2 environments.  
Note that we scale the y-axis using \textit{log} due to the huge improvements of ACL-QL on random datasets.
For instance, ACL-QL achieves 33.5 on Hopper-random-v2, which is 679.1\% times higher than the results of 4.9 for CQL-5.
As a complement, Table~\ref{tab:compare_cql_half_v2} reports the normalized results of ACL-QL and CQL in terms of the average accumulated quantitative returns on HalfCheetah environments over 6 seeds.
From Figure~\ref{fig:compare_cql_all_v2} and Table~\ref{tab:compare_cql_half_v2}, one trend we can clearly observe is that CQL with a higher conservative level $\alpha$ can usually achieve higher performance on medium-expert datasets as opposed to that a lower conservative level $\alpha$ is more effective for random datasets. 
For instance, in Table~\ref{tab:compare_cql_half_v2}, the CQL agent with $\alpha=20$ delivers a high-performance value of 85.3 on the HalfCheetah-medium-expert dataset, while dropping its performance dramatically to 6.8 on the HalfCheetah-random dataset.
The CQL agent with $\alpha=1$ drops its performance to 54.9 on the HalfCheetah-medium-expert dataset while delivering a high-performance value of 28.8 on the HalfCheetah-random dataset.
It is because a higher conservative level $\alpha$ encourages the policy to mimic the actions presented in the dataset and thus can obtain high performance on high-quality datasets.
Thus ,it is difficult for CQL to achieve satisfactory performance on all kinds of datasets with a fixed $\alpha$.
It is exactly the issue that the ACL-QL focuses on.
Since ACL-QL generates adaptive weight for each state-action pair and controls the conservative level in a more fine-grained way, ACL-QL can achieve balanced and state-of-the-art results for different dataset types.

Table~\ref{tab:half-q-comparisons} shows the comparisons of average Q-values over the datasets on HalfCheetah-v2 environments.
The Q-values of SAC~\cite{Haarnoja2018soft} are from the ordinary Q-functions, and they are larger than the Q-values of CQL and ACL-QL, showing both CQL and ACL-QL learn conservative Q-functions.
Compared to CQL-10, the Q-values of ACL-QL ($\alpha=10$ in Equation~\ref{equ:ACL-QL_theo_3_1}) are higher than CQL-10, showing ACL-QL is less conservative than CQL-10, while the Q-values also do not explode.
For medium datasets, we can observe that the average Q-values of ACL-QL is 693.8, which is between the values 1568.6 of SAC and 273.1 of CQL-10, validating the effectiveness of the losses $\mL_{ord}, \mL_{cql}$ in~\Cref{equ:ACL-QL-loss1,equ:ACL-QL-loss2}.

\subsection{Ablation Study}
\label{sec: ab_study}

\textbf{Effect of the Proposed Losses.} 
Besides the main results of the comparison to other baselines and the detailed comparisons to different conservative levels, we also conducted extensive ablation studies for ACL-QL to study the effect of the proposed losses.
We reported the quantitative results of the normalized average accumulated rewards on the HalfCheetah-v2 environment over 6 random seeds, including 4 kinds of datasets in Table~\ref{tab:abl_half_results}.

As shown in Table~\ref{tab:abl_half_results}, when we only control the conservative level using $\mL_{ord}$ (first row) or $\mL_{cql}$ (second row), ACL-QL dropped its performance seriously, with only around 38.2 and 40.3 of normalized returns even on the medium-expert dataset.
As shown in the third row, without using $\mL_{mono}$ and $\mL_{pos}$ to limit the range of the adaptive weights, it is prone to learn erroneously adaptive weights, and the errors increase like a snowball as the policy evaluation repeats and lead to failure.
From the fourth row, we can find that only using the monotonicity loss would lead to a drop in performance, such as the result of 21.5 in the medium-replay dataset.


\begin{figure*}[t]
    \begin{minipage}[t]{0.48\linewidth}
    \includegraphics[width=0.9\linewidth]{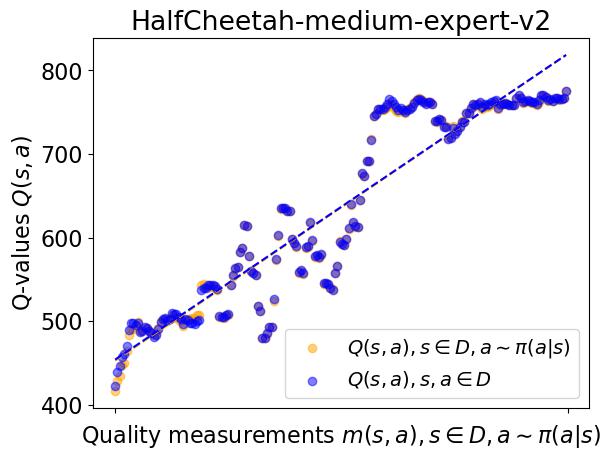}
    \end{minipage}%
    \begin{minipage}[t]{0.5\linewidth}
    \includegraphics[width=0.9\linewidth]{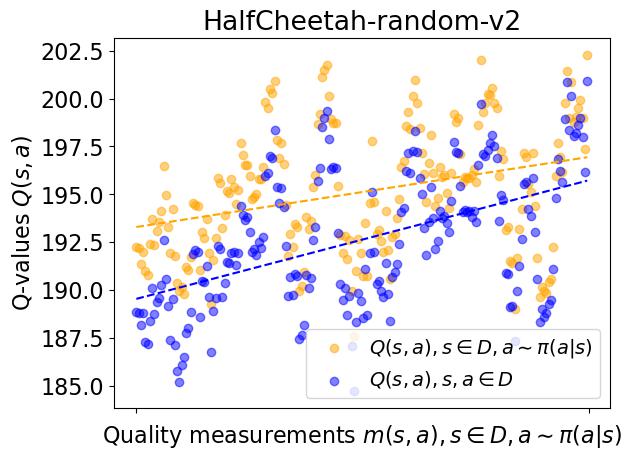}
    \end{minipage}%
\caption{Visualization of the Q-values $Q_{\mu}(s,a)$, where the actions $a$ are from the learned policy $\pi$ or dataset $D$, on HalfCheetah-medium-expert-v2 and HalfCheetah-random datasets. The Q-values are sorted in ascending order according to relative transition quality measurement $m(s,a)$. In the left figure, almost all Q-values of the predicted actions and the in-dataset actions overlap, showing the learned policy and the behavioral policy are similar. In the right figure, the Q-values of the predicted actions are higher than the Q-values of the in-dataset actions, showing ACL-QL learns a better policy.}  
\label{fig:vis_q}
\end{figure*}

\textbf{Effect of Different Transition Quality Measurements.} 
To investigate the effect of different combinations of the Monte Carlo returns and single-step rewards, we conducted extensive ablation studies for different values of $\lambda$ in~\Cref{equ:In_RTQ} and N-steps SARSA returns~\cite{Sutton2018reinforcement}.
Table~\ref{tab:abl_qua_mea_half_v2} shows the normalized average accumulated rewards of different transition quality measurements on the HalfCheetah-v2 environments over 6 random seeds.

Compared to other quality measurement choices, ACL-QL ($\lambda=0.5$) achieves the best sum results of all 4 datasets, that the sum result of ACL-QL is 241.6 while the case $\lambda=0.0$ only obtains the result of 218.9.
It presents that setting $\lambda=0.5$ is a balanced and general choice in practice because it can make a suitable trade-off between the long-term rewards and current rewards.
It is worth noting that for datasets with high quality, the performance increases as the $\lambda$ increases, which assigns more weight to the Monte Carlo returns.
For example, in the "halfcheetah-m-e-v2" task, the case of $\lambda=0.0$ achieves the result of 79.8 while the result of $\lambda=1.0$ is 93.1.
In contrast, for datasets with low quality, the performance increases as the $\lambda$ decreases, which assigns more weight to the single-step returns.
Because the Monte Carlo returns are inherently high in high-quality datasets, using higher $\lambda$ to assign more weight to the Monte Carlo returns directly contributes to training a good policy.
In the low-quality datasets, the Monte Carlo returns are usually low, but the single-step rewards may be high. 
Assigning more weight to single-step rewards with lower $\lambda$ can enable policy to "stitch" high-quality single-step action to achieve better performance.
We also evaluated using 5-step and 10-step SARSA for transition quality measurement.
N-step SARSA methods can achieve comparable results on all 4 datasets but can not excel ACL-QL.
We suggest that N-step SARSA, using the returns of the fixed-length trajectory segments, does not explicitly consider the Monte Carlo returns and single-step rewards and can only provide a limited contribution to learning an accurate Q-function.

\textbf{Effect of Different Discount Factors.} 
To investigate the role of the discount factor $\gamma$ in the offline RL problem, we set up an ablation study with the undiscounted function $\gamma=1$, which may lead to unstable policies in infinite horizon tasks (e.g., the "halfcheetah", "hopper" and "walker" tasks in D4RL)~\cite{Postoyan2016stability}.
As shown in Table~\ref{tab:abl-gamma-comparisons}, we reported the normalized average accumulated rewards on HalfCheetah-v2 environments over 6 random seeds.
We can observe that the case of $\gamma=0.99$ (i.e., our implementation) performs on par with the undiscounted function case on the "medium-expert" dataset and better on the other 3 types of datasets.
We suggest that a discounted factor (e.g., $\gamma=0.99$) can model an inaccurate estimation of future rewards and assign smaller weights to rewards that are further away from the current time step.
Therefore, it can learn a more accurate Q-function and achieve better performance, especially in the offline RL setting.
In addition, the performance of setting $\gamma=1.0$ does not lag too far behind because the trajectory length is truncated with a max horizon length (e.g., 1000 in the "halfcheetah" task) in the data collection process.
In this way, the infinite horizon tasks (e.g., the "halfcheetah" task in D4RL) are converted into finite horizon tasks in practice.

\textbf{Visualization of the Q-values.}
Besides comparing average Q-values over the dataset, we further check the Q-values of each transition in a more fine-grained way.
We randomly sampled 200 transitions and fitted a line over all transitions to show the trend of the Q-values. 
Figure~\ref{fig:vis_q} shows the visualization of the Q-values $Q_{\mu}(s,a)$, where $a$ are from the learned policy $\pi$ or dataset $D$, on HalfCheetah-medium-expert-v2 and HalfCheetah-random-v2 datasets.
The Q-values are sorted in ascending order according to the quality measurement $m(s,a)$.

For the HalfCheetah-medium-expert-v2 dataset shown on the left in Figure~\ref{fig:vis_q}, we can observe that almost all Q-values of the predicted actions ($s \in D, a \sim \pi(a|s)$, yellow points) and the in-dataset actions ($s, a \in D$, blue points) are overlapped.
This represents a very close match between the learned policy $\pi$ and the behavioral policy $\pi_{\beta}$, resulting from the high-quality dataset (i.e., ACL-QL is trained to mimic the behavioral policy).

For the HalfCheetah-random-v2 dataset shown on the right in Figure~\ref{fig:vis_q}, we can observe that almost all Q-values of the predicted actions ($s \in D, a \sim \pi(a|s)$, yellow points) are above the Q-values of in-dataset actions ($s, a \in D$, blue points).
This represents ACL-QL learns a policy different from the behavioral policy and can choose better actions than the behavioral policy.
We can also observe that for both datasets, the fitted curves are monotonically increasing, showing the Q-values of each transition Q(s, a), whether the action $a$ is from the learned policy $\pi$ or dataset,  is proportional to the quality measurements $m(s, a)$.

\section{Conclusion}

In this paper, we proposed a flexible framework named Adaptive Conservative Level in Q-Learning (ACL-QL), which sheds light on how to control the conservative level of the Q-function in a fine-grained way.
In the ACL-QL framework, two weight functions corresponding to the out-of-distribution (OOD) actions and actions in the dataset are introduced to adaptively shape the Q-function. 
More importantly, the form of these two adaptive weight functions is not fixed and it is possible to define particular forms for different scenarios, e.g., elaborately hand-designed rules or learnable deep neural networks. 
We provide a detailed theoretical analysis of how the conservative level of the learned Q-function changes under different conditions and define the monotonicity of the adaptive weight functions.
To illustrate the feasibility of our framework, we propose a novel practical algorithm ACL-QL using neural networks as the weight functions.
With the guidance of the theoretical analysis, we construct two surrogate and monotonicity losses to control the conservative level and maintain the monotonicity. 
We build extensive experiments on commonly-used offline RL benchmarks and the state-of-the-art results well demonstrate the effectiveness of our method.

\ifCLASSOPTIONcaptionsoff
  \newpage
\fi

\bibliographystyle{IEEEtran}
\bibliography{reference}

\newpage

\appendix

\section{Proofs}
\label{appendix:A}

\setcounter{equation}{25}

\textbf{Proof of Proposition IV.1}.
Without considering the sampling error between the empirical $\hatB^{\pi} \hatQ$ and ordinary Bellman backups $\mB^{\pi} \hatQ$,
we first show the optimization problem of Q-function in ACL-QL as the following:
\begin{align}
\label{supp_equ:ACL-QL_Q_problem}
    \hatQ^{k+1}_{ACL-QL} \leftarrow \rmmin \limits_{Q} 
    \left(
    \bbE_{s \sim \mD, a \sim \mu(a|s)} \left[ w_{\mu}(s,a) \cdot Q(s,a) \right] \right. \notag
    \\
    \left. - \bbE_{s \sim \mD, a \sim \pi_{\beta}(a|s)} \left[ w_{\BH}(s,a) \cdot Q(s,a) \right]
    \right) \notag
    \\
    + \frac{1}{2} \bbE_{s,a,\nexts \sim \mD}
    \left[
    \left( Q(s,a) - \mB^{\pi} \hatQ^{k}(s,a) \right)^2
    \right]
\end{align}

By setting the derivative of~\Cref{supp_equ:ACL-QL_Q_problem} to 0, we can obtain the form of the resulting Q-function $\hatQ^{k+1}$ in ACL-QL:
\begin{align}
    & \forall s \in \mD, a, \frac{\partial \hatQ^{k+1}_{ACL-QL}}{\partial Q} = 0 \\
    \Rightarrow \ & w_{\mu} \cdot \mu - w_{\BH} \cdot \pi_{\beta} + \pi_{\beta} \cdot (Q - \mB^{\pi} \hatQ^{k}) = 0 \\
    \Rightarrow \ & \hatQ^{k+1}_{ACL-QL} = \mB^{\pi} \hatQ^{k} - \frac{w_{\mu} \cdot \mu - w_{\BH} \cdot \pi_{\beta}}{\pi_{\beta}} \label{supp_equ:ACL-QL_Q-deri}.
\end{align}

Note that the ordinary Q-function is only derived from the ordinary Bellman operator $Q^{k+1} = \mB^{\pi} \hatQ^{k}$.
Based on the~\Cref{supp_equ:ACL-QL_Q-deri}, we have exactly the condition to control the conservative level over the Q-function, as shown in Equation 5.
If we want to relax the conservative level, like requiring control over the V-values or the empirical MDP, we can also easily relax the~\Cref{supp_equ:ACL-QL_Q-deri} to the integration over each state or the whole empirical MDP respectively.
For the conditions where we want to make the learned Q-function $\hatQ^{\pi}_{ACL-QL}$ is less conservative than the ordinary Q-function $Q^{\pi}$, we can easily replace the "$>$" to "$<$" in Proposition IV.1

Next, we show that ACL-QL bounds the gap between the learned Q-values and ordinary Q-values with the consideration of the sampling error between the empirical $\hatB^{\pi} \hatQ$ and actual Bellman operator $\mB^{\pi} \hatQ$.
Following~\cite{Auer2008near,Osband2016deep,Kumar2020conservative}, the error can be bounded by leveraging the concentration properties of $\hatB^{\pi}$.
We introduce the bound in brief here:
with high probability $\geq 1 - \delta, | \hatB^{\pi} \hatQ - \mB^{\pi} \hatQ|(s,a) \leq \frac{C_{r, \mP, \delta}}{\sqrt{|\mD (s,a)|}}, \forall s, a \in \mD$, where $C_{r, \mP, \delta}$ is a constant relating to the reward function $r (s,a)$, environment dynamic $\mP (\cdot |s,a)$, and $\delta \in (0.1)$.

\textbf{Proof of Proposition VI.2}.
In Proposition IV.1, 
we calculate the gap between the learned Q-function $\hatQ^{\pi}_{ACL-QL}$ and the ordinary Q-function $Q^{\pi}$, representing the conservative level, without the sampling error between the empirical $\hatB^{\pi} \hatQ$ and ordinary Bellman backups $\mB^{\pi} \hatQ$.
Now, we can obtain a more precise bound for the conservative level with the consideration of the sampling error.
Following~\cite{Auer2008near,Osband2016deep,Kumar2020conservative}, we can relate the  empirical Bellman backups $\hatB^{\pi} \hatQ$ and ordinary Bellman backups $\mB^{\pi} \hatQ$ as the following:
with high probability $\geq 1 - \delta, \delta \in (0, 1)$,
\begin{align}
\label{supp_equ:ACL-QL_sampling-err}
    & \forall Q, s, a \in \mD, \notag \\ 
    & \ |\hatB^{\pi} \hatQ - \mB^{\pi} \hatQ|(s,a) \leq \frac{C_{r, \mP, \delta} R_{max}}{(1 - \gamma) \sqrt{|\mD (s,a)|}}, 
\end{align}
where $R_{max}$ is the upper bound for the reward function (i.e., $|r (s,a)| \leq R_{max}$), $C_{r, \mP, \delta}$ is a constant relating to the reward function $r (s,a)$, environment dynamic $\mP (\cdot |s,a)$.
More detailed proofs of the relationship are provided in~\cite{Kumar2020conservative}.

For the right inequality in Equation 6,
we reason the fixed point of the Q-function in ACL-QL as the following:
\begin{align}
\label{supp_equ:ACL-QL_theory_2_right}
    & |\hatB^{\pi} \hatQ_{ACL-QL} - \mB^{\pi} \hatQ_{ACL-QL}|(s,a) \leq \frac{C_{r, \mP, \delta} R_{max}}{(1 - \gamma) \sqrt{|\mD (s,a)|}} \\
    \Rightarrow \ & \hatB^{\pi} \hatQ_{ACL-QL} \leq  \mB^{\pi} \hatQ_{ACL-QL} + \frac{C_{r, \mP, \delta} R_{max}}{(1 - \gamma) \sqrt{|\mD (s,a)|}} \\
    \Rightarrow \ & \hatB^{\pi} \hatQ_{ACL-QL} 
    - \frac{w_{\mu} \cdot \mu - w_{\BH} \cdot \hatP_{\beta}}{\hatP_{\beta}} 
    \leq \mB^{\pi} \hatQ_{ACL-QL} \notag \\
    & - \frac{w_{\mu} \cdot \mu - w_{\BH} \cdot \hatP_{\beta}}{\hatP_{\beta}} + \frac{C_{r, \mP, \delta} R_{max}}{(1 - \gamma) \sqrt{|\mD (s,a)|}} \\
    \Rightarrow \ & \hatQ^{\pi}_{ACL-QL}
    \leq 
    \mB^{\pi} \hatQ_{ACL-QL} - \frac{w_{\mu} \cdot \mu - w_{\BH} \cdot \hatP_{\beta}}{\hatP_{\beta}} \notag \\
    & + \frac{C_{r, \mP, \delta} R_{max}}{(1 - \gamma) \sqrt{|\mD (s,a)|}} \\
    \Rightarrow \ & \hatQ^{\pi}_{ACL-QL} 
    \leq
    (r + \gamma P^{\pi} \hatQ^{\pi}_{ACL-QL} ) - \frac{w_{\mu} \cdot \mu - w_{\BH} \cdot \hatP_{\beta}}{\hatP_{\beta}} \notag \\
    & + \frac{C_{r, \mP, \delta} R_{max}}{(1 - \gamma) \sqrt{|\mD (s,a)|}} \\
    \Rightarrow \ & \hatQ^{\pi}_{ACL-QL} 
    \leq
    (I - \gamma P^{\pi})^{-1} \left [ r - \frac{w_{\mu} \cdot \mu - w_{\BH} \cdot \hatP_{\beta}}{\hatP_{\beta}} \right. \notag \\
    & \left. + \frac{C_{r, \mP, \delta} R_{max}}{(1 - \gamma) \sqrt{|\mD (s,a)|}} \right ] \\
    \Rightarrow \ & \hatQ^{\pi}_{ACL-QL}
    \leq
    Q^{\pi} -  (I - \gamma P^{\pi})^{-1} \frac{w_{\mu} \cdot \mu - w_{\BH} \cdot \hatP_{\beta}}{\hatP_{\beta}} \notag \\
    & + (I - \gamma P^{\pi})^{-1} \frac{C_{r, \mP, \delta} R_{max}}{(1 - \gamma) \sqrt{|\mD (s,a)|}} \\
    \Rightarrow \ & \hatQ^{\pi}_{ACL-QL} - Q^{\pi} 
    \leq
    g(s,a) + err(s,a).
\end{align}

For the left inequality in Equation 7,
we have a similar process as the following:
\begin{align}
\label{supp_equ:ACL-QL_theory_2_left}
    & |\hatB^{\pi} \hatQ_{ACL-QL} - \mB^{\pi} \hatQ_{ACL-QL}|(s,a) \leq \frac{C_{r, \mP, \delta} R_{max}}{(1 - \gamma) \sqrt{|\mD (s,a)|}} \\
    \Rightarrow \ & \hatB^{\pi} \hatQ_{ACL-QL} \geq  \mB^{\pi} \hatQ_{ACL-QL} - \frac{C_{r, \mP, \delta} R_{max}}{(1 - \gamma) \sqrt{|\mD (s,a)|}} \\
    \Rightarrow \ & \hatB^{\pi} \hatQ_{ACL-QL} - \frac{w_{\mu} \cdot \mu - w_{\BH} \cdot \hatP_{\beta}}{\hatP_{\beta}} 
    \geq 
    \mB^{\pi} \hatQ_{ACL-QL} \notag \\
    & - \frac{w_{\mu} \cdot \mu - w_{\BH} \cdot \hatP_{\beta}}{\hatP_{\beta}} - \frac{C_{r, \mP, \delta} R_{max}}{(1 - \gamma) \sqrt{|\mD (s,a)|}} \\
    \Rightarrow \ & \hatQ^{\pi}_{ACL-QL}
    \geq 
    \mB^{\pi} \hatQ_{ACL-QL} - \frac{w_{\mu} \cdot \mu - w_{\BH} \cdot \hatP_{\beta}}{\hatP_{\beta}} \notag \\
    & - \frac{C_{r, \mP, \delta} R_{max}}{(1 - \gamma) \sqrt{|\mD (s,a)|}} \\
    \Rightarrow \ & \hatQ^{\pi}_{ACL-QL} 
    \geq
    (r + \gamma P^{\pi} \hatQ^{\pi}_{ACL-QL} ) - \frac{w_{\mu} \cdot \mu - w_{\BH} \cdot \hatP_{\beta}}{\hatP_{\beta}} \notag \\
    & - \frac{C_{r, \mP, \delta} R_{max}}{(1 - \gamma) \sqrt{|\mD (s,a)|}} \\
    \Rightarrow \ & \hatQ^{\pi}_{ACL-QL} 
    \geq
    (I - \gamma P^{\pi})^{-1} \left [ r - \frac{w_{\mu} \cdot \mu - w_{\BH} \cdot \hatP_{\beta}}{\hatP_{\beta}} \right. \notag \\
    & \left. - \frac{C_{r, \mP, \delta} R_{max}}{(1 - \gamma) \sqrt{|\mD (s,a)|}} \right ] \\
    \Rightarrow \ & \hatQ^{\pi}_{ACL-QL} 
    \geq
    Q^{\pi} -  (I - \gamma P^{\pi})^{-1} \frac{w_{\mu} \cdot \mu - w_{\BH} \cdot \hatP_{\beta}}{\hatP_{\beta}} \notag \\
    & -  (I - \gamma P^{\pi})^{-1} \frac{C_{r, \mP, \delta} R_{max}}{(1 - \gamma) \sqrt{|\mD (s,a)|}} \\
    \Rightarrow \ & \hatQ^{\pi}_{ACL-QL} - Q^{\pi} 
    \geq
    g(s,a) - err(s,a).
\end{align}

\textbf{Proof of Proposition IV.3}.
As shown in CQL~\cite{Kumar2020conservative}, we first recap the optimization problem of CQL as the following:
\begin{align}
\label{supp_equ:CQL_Q_problem}
    \hatQ^{k+1}_{CQL} \leftarrow \rmmin \limits_{Q} \alpha 
    \left(
    \bbE_{s \sim \mD, a \sim \mu(a|s)} \left[ Q(s,a) \right] \right. \notag
    \\
    \left. - \bbE_{s \sim \mD, a \sim \pi_{\beta}(a|s)} \left[ Q(s,a) \right]
    \right) \notag
    \\
    + \frac{1}{2} \bbE_{s,a,\nexts \sim \mD}
    \left[
    \left( Q(s,a) - \mB^{\pi} \hatQ^{k}(s,a) \right)^2
    \right].
\end{align}
We can observe that CQL is the a special case of ACL-QL where the adaptive weight functions $w_{\mu}(s,a)$ and $w_{\BH}(s,a)$ are both constant $\alpha$.
By setting the derivative of~\Cref{supp_equ:CQL_Q_problem} to 0, we can obtain the form of the resulting Q-function $\hatQ^{k+1}$ in CQL:
\begin{align}
    & \forall s \in \mD, a, \frac{\partial \hatQ^{k+1}_{CQL}}{\partial Q} = 0 \\
    \Rightarrow \ & \alpha \cdot \mu - \alpha \cdot \pi_{\beta} + \pi_{\beta} \cdot (Q - \mB^{\pi} \hatQ^{k}) = 0 \\
    \Rightarrow \ & \hatQ^{k+1}_{CQL} = \mB^{\pi} \hatQ^{k} - \alpha \frac{\mu - \pi_{\beta}}{\pi_{\beta}} \label{supp_equ:CQL_Q-deri}.
\end{align}

Similar to the proof of Proposition IV.1,
we calculate the difference between the Q-values of ACL-QL and CQL in~\Cref{supp_equ:ACL-QL_Q-deri,supp_equ:CQL_Q-deri} as the following:
\begin{align}
    & \forall s \in \mD, a, \ \hatQ^{k+1}_{ACL-QL} - \hatQ^{k+1}_{CQL} \\
    & = \mB^{\pi} \hatQ^{k} - \frac{w_{\mu} \cdot \mu - w_{\BH} \cdot \pi_{\beta}}{\pi_{\beta}} - \mB^{\pi} \hatQ^{k} + \alpha \frac{\mu - \pi_{\beta}}{\pi_{\beta}} \\
    & = \frac{(\alpha - w_{\mu}) \mu - (\alpha - w_{\BH}) \pi_{\beta}}{\pi_{\beta}}. \label{supp_equ:ACL-QL_CQL_Q-diff}
\end{align}
If we want to learn a less conservative Q-function than CQL in ACL-QL, we need to make the difference in~\Cref{supp_equ:ACL-QL_CQL_Q-diff} greater than 0, as shown in Equation 10.
If we want to relax the conservative level like requiring control over the V-values or the empirical MDP, we can also easily relax the~\Cref{supp_equ:ACL-QL_CQL_Q-diff} to the integration over each state or the whole empirical MDP respectively.
For the conditions where we want to make the learned Q-function $\hatQ^{\pi}_{ACL-QL}$ is more conservative than the CQL Q-function $\hatQ^{\pi}_{CQL}$, we can easily replace the "$\geq$" to "$\leq$" in Proposition IV.3

\textbf{Proof of Lemma V.1}.

Suppose $f(x) = \ln{x} - x + 1$, then $f'(x) = \frac{1-x}{x}$.
It is easy to know that $f(x) \uparrow$ over $(0,1)$ and $f(x) \downarrow$ over $(1,+ \infty)$.
Then $f(x) \leq f(1) = 0$.

\section{From theoretical framework to algorithmic implementation}

\dif{
In this section, we provide additional insights and details on how to derive the practical implementation of the algorithm from the theoretical framework of ACL-QL. 
This includes elucidating the connection between the formulation of Equations (15), (16), (17), and (23) and the framework discussed in Section IV.

\subsection{Interpretation of Equation 15}

Recall that Definition IV.1 in Section IV depicts the property that a good action should have a higher Q-value than a bad action and defines the monotonicity of the adaptive weight functions as follows:

Definition IV.1
(The monotonicity of the adaptive weight functions). For any state $s \in \mD$, the monotonicity of the adaptive weight functions is defined as that a good action $a \in \mA$ with a higher ordinary Q value has a lower $w_{\mu}$ value and higher $w_{\BH}$ value:
\begin{align}
\label{supp_equ:ACL-QL_defi_1}
    & \forall s_i, s_j \in \mD, a_i, a_j \in \mu(a|s), \notag \\
    & \quad w_{\mu} (s_i, a_i) - w_{\mu} (s_j, a_j)  \propto  Q^{*}(s_j, a_j) - Q^{*}(s_i, a_i). \tag{11} \\
    & \forall s_i, s_j \in \mD, a_i, a_j \in \hatP_{\beta}(a|s), \notag \\
    & \quad w_{\BH} (s_i, a_i) - w_{\BH} (s_j, a_j)  \propto  Q^{*}(s_i, a_i) - Q^{*}(s_j, a_j), \tag{12}
\end{align}

where $Q^{*}$ is the optimal Q-function.
In Definition IV.1, we use the optimal Q-function which is a natural ideal metric to measure the action quality to define the monotonicity.
However, it is an ill-posed problem since if we know the optimal Q-function, we can directly train an optimal policy and solve the problem. 

Therefore, to make the above equations optimizable, we must first identify a suitable and computable method to replace the $Q^{*}$ function. 
In Section V, we proposed Equations (13) and (14) to estimate the quality of both in-distribution and out-of-distribution action data points. 
Now, we can replace the $Q^{*}$ function with the quality function, yielding the following equation: 
\begin{align}
\label{supp_equ:ACL-QL_defi_step_1}
    & \forall s_i, s_j \in \mD, a_i, a_j \in \mu(a|s), \notag \\
    & \quad w_{\mu} (s_i, a_i) - w_{\mu} (s_j, a_j)  \propto  m(s_j, a_j) - m(s_i, a_i). \\
    & \forall s_i, s_j \in \mD, a_i, a_j \in \hatP_{\beta}(a|s), \notag \\
    & \quad w_{\BH} (s_i, a_i) - w_{\BH} (s_j, a_j)  \propto  m(s_i, a_i) - m(s_j, a_j),
\end{align}

However, in Equations (54) and (55), the optimal ratio between the difference in the quality function and the difference in the adaptive coefficient is unclear. 
To address this, we applied the softmax function to constrain both within the range (0, 1) and replaced the proportionality sign with an equality sign. 
This leads to the following equation:

\begin{align}
\label{supp_equ:ACL-QL_defi_step_2}
    & \forall s_i, s_j \in \mD, a_i, a_j \in \mu(a|s), \notag \\
    & \quad \sigma(w_{\mu} (s_i, a_i)) - \sigma(w_{\mu} (s_j, a_j))  =  \sigma(m(s_j, a_j)) \notag \\ 
    & \qquad \qquad \qquad \qquad \qquad \qquad - \sigma(m(s_i, a_i)). \\
    & \forall s_i, s_j \in \mD, a_i, a_j \in \hatP_{\beta}(a|s), \notag \\
    & \quad \sigma(w_{\BH} (s_i, a_i)) - \sigma(w_{\BH} (s_j, a_j))  =  \sigma(m(s_i, a_i)) \notag \\
    & \qquad \qquad \qquad \qquad \qquad \qquad - \sigma(m(s_j, a_j)),
\end{align}

Then, we use Mean Squared Error (MSE) for computation efficiency and simplicity.
The monotonicity loss in Equation (15) is defined as the following:

\begin{align}
\label{supp_equ:ACL-QL-mono-loss}
    & \forall (s_i, a_i), (s_j, a_j) \notin \mD, (s_k, a_k), (s_l, a_l) \in \mD, \notag
    \\ 
    & \mL_{mono}(w_{\mu}, w_{\BH}) = 
    \lVert 
    \left( \sigma(w_{\mu} (s_i, a_i)) - \sigma(w_{\mu} (s_j, a_j)) \right) \notag
    \\
    & \qquad \qquad \qquad - \left(\sigma(m(s_j, a_j)) - \sigma(m(s_i, a_i)) \right)
    \rVert_{2}^{2} \notag
    \\
    & \qquad \qquad \qquad + \lVert 
    \left( \sigma(w_{\BH} (s_k, a_k)) - \sigma(w_{\BH} (s_l, a_l)) \right) \notag
    \\
    & \qquad \qquad \qquad - \left(\sigma(m(s_k, a_k)) - \sigma(m(s_l, a_l)) \right)
    \rVert_{2}^{2}, \tag{15}
\end{align}

\subsection{Interpretation of Equation 16}

Recall that the Proposition IV.1 is as follows:

Proposition IV.1
(The conservative level of ACL-QL).
For any $\mu$ with $\rmsupp \mu \subset \rmsupp \hatP_{\beta}$, without considering the sampling error between the empirical $\hatB^{\pi} \hatQ$ and ordinary Bellman backups $\mB^{\pi} \hatQ$, the conservative level of ACL-QL can be controlled over the Q-values. 
The difference $d_{ord} (s,a)$ between the ordinary Q-function $Q^{\pi}$ and the learned Q-function $\hatQ^{\pi}$ is
\begin{align}
\label{supp_equ:ACL-QL_theo_1_1}
    \forall s \in \mD, d_{ord} (s,a) & = Q^{\pi} - \hatQ^{\pi} \notag \\
    & =  \frac{ w_{\mu} \cdot \mu - w_{\BH} \cdot \pi_{\beta}}{\pi_{\beta}}. \tag{5}
\end{align}

Note that when $d_{ord} (s,a) > 0$, the learned Q-function $\hatQ^{\pi}$ is more conservative than the ordinary Q-function $Q^{\pi}$ point-wise, and it aligns with our objective.
We aim to optimize $w_{\mu}$ and $w_{\BH}$ so that the following equation holds:
\begin{align}
\label{supp_equ:ACL-QL_theo_1_1_step_1}
    \frac{ w_{\mu} \cdot \mu - w_{\BH} \cdot \pi_{\beta}}{\pi_{\beta}} > 0. 
\end{align}

However, the current equation~\ref{supp_equ:ACL-QL_theo_1_1_step_1} does not specify how much more conservative the learned Q-function is compared to the ordinary Q-function. 
To address this, we replace the zero on the right-hand side of the inequality with a distance function $d_{ord} (s,a)$, which can be implemented in any form according to the user's needs. 
This allows for flexible control over how much lower the learned Q-values are compared to the ordinary Q-values, as long as the minimum value of $d_{ord} (s,a)$ remains greater than 0. 
The further optimized objective is as follows:

\begin{align}
\label{supp_equ:ACL-QL_theo_1_1_step_2}
    \frac{ w_{\mu} \cdot \mu - w_{\BH} \cdot \pi_{\beta}}{\pi_{\beta}} > d_{ord} (s,a). 
\end{align}

Yet, the specific form of the function $d_{ord} (s,a)$ is still unknown in the above Equation~\ref{supp_equ:ACL-QL_theo_1_1_step_2}. 
Therefore, in Equation 18 of Section V, we designed the function $d_{ord} (s,a)$, substituted it into the above Equation~\ref{supp_equ:ACL-QL_theo_1_1_step_2}, and now the optimized objective is as follows:

\begin{align}
\label{supp_equ:ACL-QL_theo_1_1_step_3}
    w_{\mu} \cdot \mu - w_{\BH} \cdot \pi_{\beta} - d_{ord} (s,a) \cdot \pi_{\beta} > 0. 
\end{align}

Then, we use hinge loss to optimize the objective, ultimately leading to Equation 16: 

\begin{align}
    \mL_{ord}(w_{\mu}, w_{\BH}) & = max(0, w_{\BH} \cdot \BH \notag 
    \\
    & - w_{\mu} \cdot \mu + d_{ord} \cdot \BH), \tag{16} \label{supp_equ:ACL-QL-loss1}
\end{align}

\subsection{Interpretation of Equation 17}

Recall that the Proposition IV.3 is as follows:

Proposition IV.3
(The conservative level compared to CQL).
For any $\mu$ with $\rmsupp \mu \subset \rmsupp \hatP_{\beta}$,
given the Q-function learned from CQL is $\hatQ^{\pi}_{CQL}(s,a) = Q^{\pi} - \alpha \frac{\mu - \pi_{\beta}}{\pi_{\beta}}$, 
similar to Proposition IV.1, the conservative level of ACL-QL compared to CQL can be controlled over the Q-values.
The difference $d_{cql}(s,a)$ between the learned Q-function $\hatQ^{\pi}$ and the CQL Q-function $\hatQ^{\pi}_{CQL}$ is
\begin{align}
\label{equ:ACL-QL_theo_3_1}
    \forall s \in \mD, d_{cql}(s,a) & = \hatQ^{\pi} - \hatQ^{\pi}_{CQL} \notag \\ 
    & = \frac{(\alpha - w_{\mu}) \mu - (\alpha - w_{\BH}) \pi_{\beta}}{\pi_{\beta}}. \tag{10}
\end{align}

When $d_{cql} (s,a) > 0$, the learned Q-function $\hatQ^{\pi}$ is less conservative than the CQL Q-function $\hatQ^{\pi}_{CQL}$ point-wise.
We aim to optimize $w_{\mu}$ and $w_{\BH}$ so that the following equation holds:
\begin{align}
\label{supp_equ:ACL-QL_theo_3_1_step_1}
    \frac{(\alpha - w_{\mu}) \mu - (\alpha - w_{\BH}) \pi_{\beta}}{\pi_{\beta}} > 0. 
\end{align}

However, the current equation~\ref{supp_equ:ACL-QL_theo_3_1_step_1} does not specify how much less conservative the learned Q-function is compared to the CQL Q-function. 
To address this, we replace the zero on the right-hand side of the inequality with a distance function $d_{cql} (s,a)$, which can be implemented in any form according to the user's needs. 
This allows for flexible control over how much lower the learned Q-values are compared to the CQL Q-values, as long as the minimum value of $d_{cql} (s,a)$ remains greater than 0. 
The further optimized objective is as follows:
\begin{align}
\label{supp_equ:ACL-QL_theo_3_1_step_2}
    \frac{(\alpha - w_{\mu}) \mu - (\alpha - w_{\BH}) \pi_{\beta}}{\pi_{\beta}} > d_{cql} (s,a). 
\end{align}

Yet, the specific form of the function $d_{cl} (s,a)$ is still unknown in the above Equation~\ref{supp_equ:ACL-QL_theo_3_1_step_2}. 
Therefore, in Equation 19 of Section V, we designed the function $d_{cql} (s,a)$, substituted it into the above Equation~\ref{supp_equ:ACL-QL_theo_3_1_step_2}, and now the optimized objective is as follows:

\begin{align}
\label{supp_equ:ACL-QL_theo_1_1_step_3}
    (\alpha - w_{\mu}) \cdot \mu - (\alpha - w_{\BH}) \cdot \pi_{\beta} - d_{cql} \cdot \BH > 0. 
\end{align}

Then, we use hinge loss to optimize the objective, ultimately leading to Equation 17: 

\begin{align}
    \mL_{cql}(w_{\mu}, w_{\BH}) & = max(0, (w_{\mu} - \alpha) \cdot \mu \notag
    \\
    & - (w_{\BH} - \alpha) \cdot \BH + d_{cql} \cdot \BH), \tag{17 }\label{equ:ACL-QL-loss2} 
\end{align}

\subsection{Interpretation of Equation 23}

The family of optimization problems of our ACL-QL framework is presented below:

\begin{align}
\label{supp_equ:ACL-QL}
    \rmmin \limits_{Q} & \rmmax \limits_{\mu} 
    \left(
    \bbE_{s \sim \mD, a \sim \mu(a|s)} \left[ w_{\mu}(s,a) \cdot Q(s,a) \right] \right.
    \notag
    \\
    & \left. - \bbE_{s \sim \mD, a \sim \hatP_{\beta}(a|s)} \left[ w_{\BH}(s,a) \cdot Q(s,a) \right] 
    \right) \tag{4}
    \\
    & + \frac{1}{2} \bbE_{s,a,\nexts \sim \mD}
    \left[
    \left( Q(s,a) - \hatB^{\pi} \hatQ^{k}(s,a)  \right)^2
    \right] + \mR(\mu). \notag
\end{align}

Recall that offline RL is very challenging due to the distribution shift between the behavioral policy and the learned policy over the training process.
It often leads to the overestimation of values of out-of-distribution (OOD) actions and thus misleads the policy into choosing these erroneously estimated actions.
While the target Q-values are calculated using OOD actions and have erroneously high values, the policy is then optimized towards predicting OOD actions. 
This error then propagates and intensifies with bootstrapping during training, eventually leading to the explosion and overflow of the Q-values.

We want to suppress the Q-values of OOD actions and lift the Q-values of in-dataset actions into a moderate range to avoid the problem of overestimation while avoiding the problem of being too conservative.
To learn a conservative Q-function, the adaptive weight should be positive, and thus, we add a regularization term as follows:
\begin{align}
    \mL_{pos}(w_{\mu}, w_{\BH}) & = max(0, - w_{\mu}) + max(0,  - w_{\BH}). \tag{23} \label{equ:ACL-QL-pos_loss}
\end{align}

}

\section*{Biographies}

\begin{IEEEbiography}[{\includegraphics[width=1in,height=1.25in, clip, keepaspectratio]{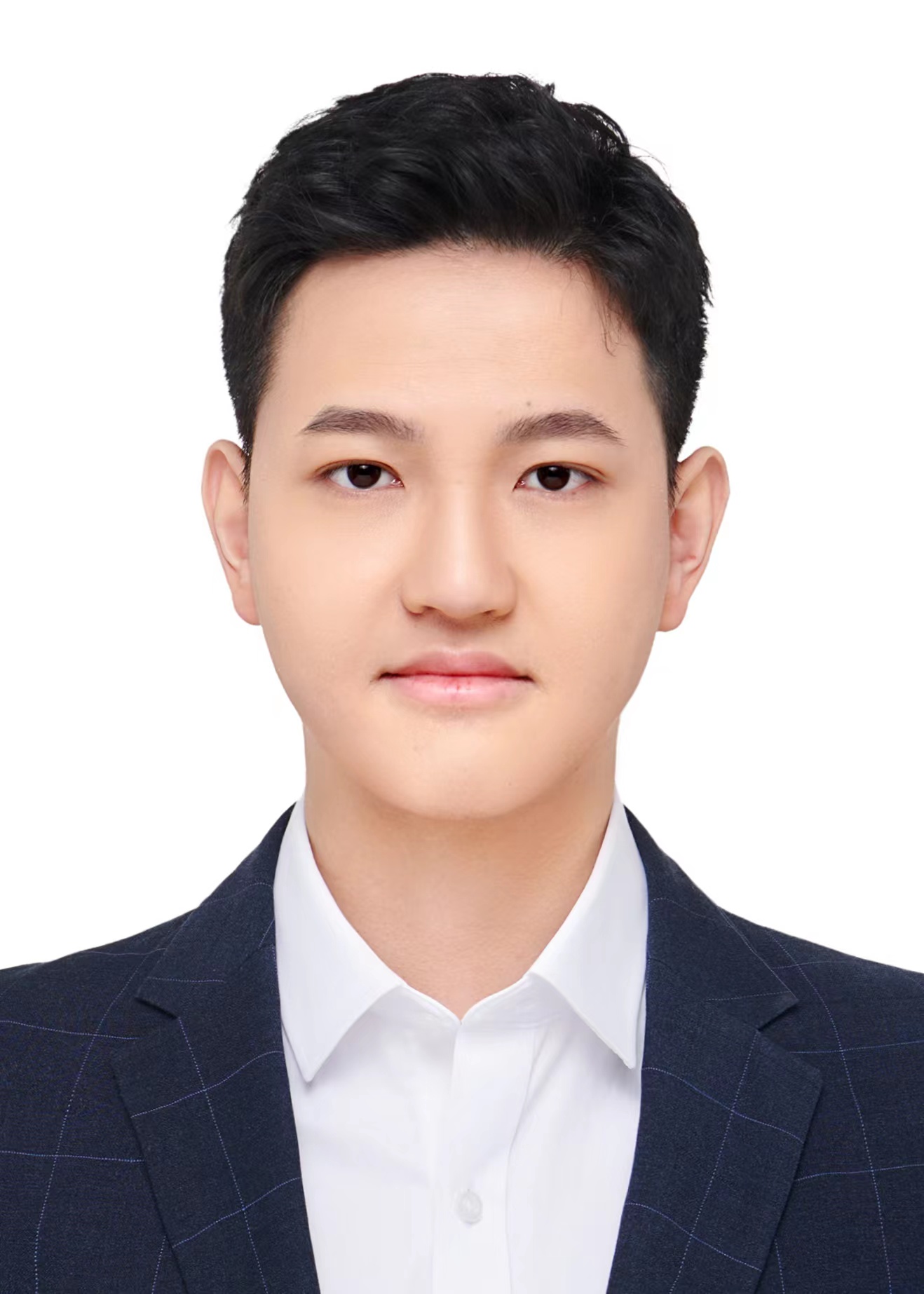}}]{Kun Wu}
received his B.S. degree from Beijing Institute of Technology in 2017. He is currently working towards his Ph.D degree in the Department of Electrical Engineering and Computer Science at Syracuse University. His research interests include Computer Vision, Reinforcement Learning, and Robot Learning.
\end{IEEEbiography}

\begin{IEEEbiography}[{\includegraphics[width=1in,height=1.25in, clip, keepaspectratio]{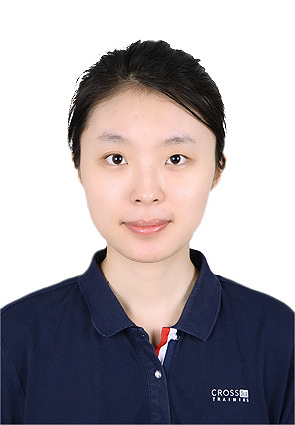}}]{Yinuo Zhao}
received her B.S. degree from Beijing Institute of Technology in 2019. She is currently working towards her Ph.D degree in the School of Computer Science at Beijing Institute of Technology. Her research interests lie in the area of generative models, deep reinforcement learning, and robot manipulation.
\end{IEEEbiography}

\begin{IEEEbiography}[{\includegraphics[width=1in,height=1.25in, clip, keepaspectratio]{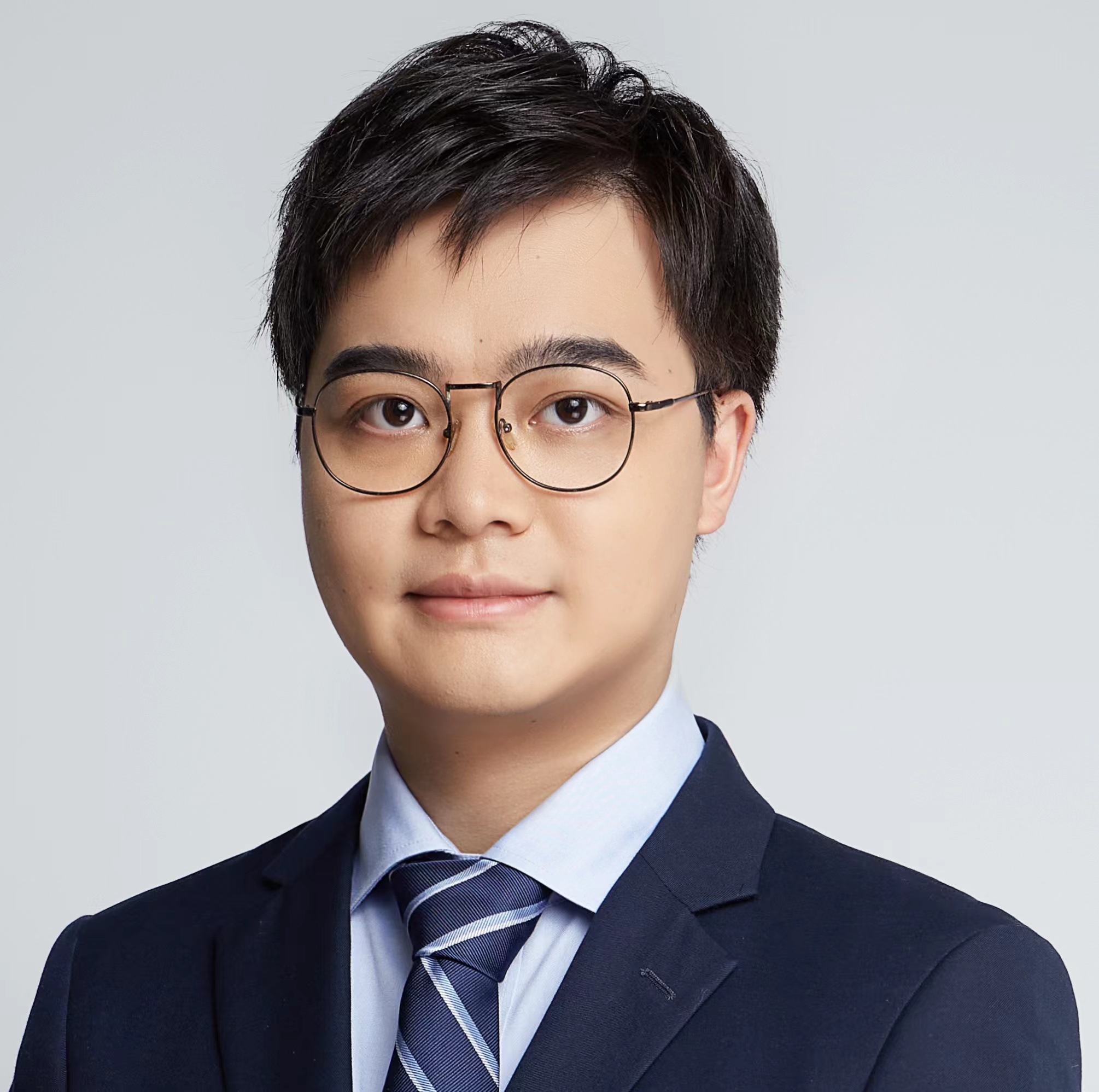}}]{Zhiyuan Xu}
received his Ph.D. degree in computer information science and engineering from Syracuse University, Syracuse, NY, USA, in 2021, and his B.E. degree in the School of Computer Science and Engineering from the University of Electronic Science and Technology of China, Chengdu, China, in 2015. He is with Beijing Innovation Center of Humanoid Robotics, Beijing, China. His research interests include Deep Learning, Deep Reinforcement Learning, and Robot Learning.
\end{IEEEbiography}

\begin{IEEEbiography}[{\includegraphics[width=1in,height=1.25in, clip, keepaspectratio]{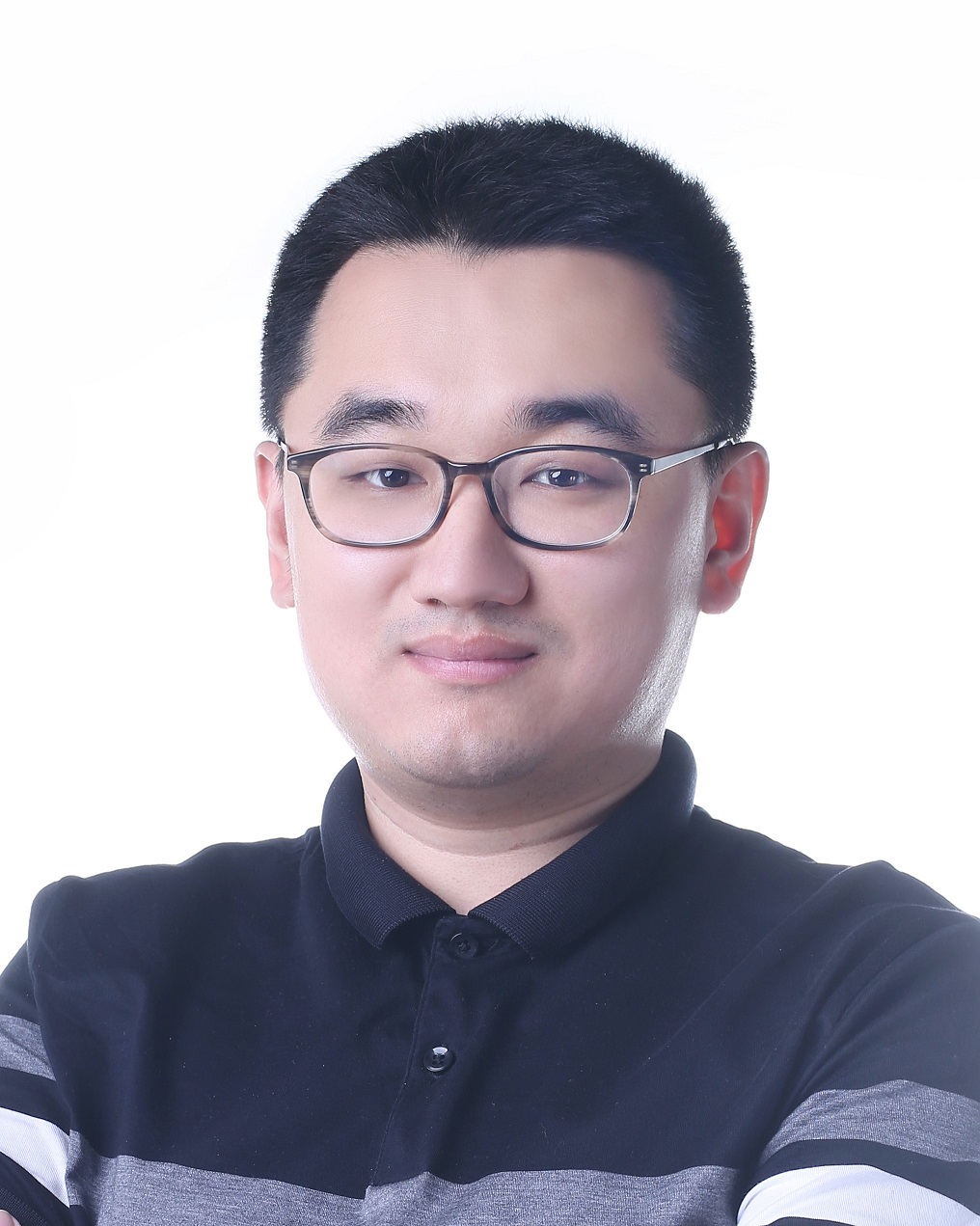}}]{Zhengping Che}
(Member, IEEE) received the Ph.D. degree in Computer Science from the University of Southern California, Los Angeles, CA, USA, in 2018, and the B.E. degree in Computer Science (Yao Class) from Tsinghua University, Beijing, China, in 2013. 
He is with Beijing Innovation Center of Humanoid Robotics, Beijing, China.
His current research interests lie in embodied AI and deep learning.
\end{IEEEbiography}

\begin{IEEEbiography}[{\includegraphics[width=1in,height=1.25in, clip, keepaspectratio]{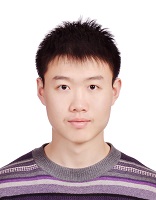}}]{Chengxiang Yin}
received his B.S. degree from Beijing Institute of Technology in 2016. He is currently working toward his Ph.D degree in the Department of Electrical Engineering and Computer Science at Syracuse University. His research interests include Machine Learning and Computer Vision.
\end{IEEEbiography}

\begin{IEEEbiography}[{\includegraphics[width=1in,height=1.25in,clip,keepaspectratio]{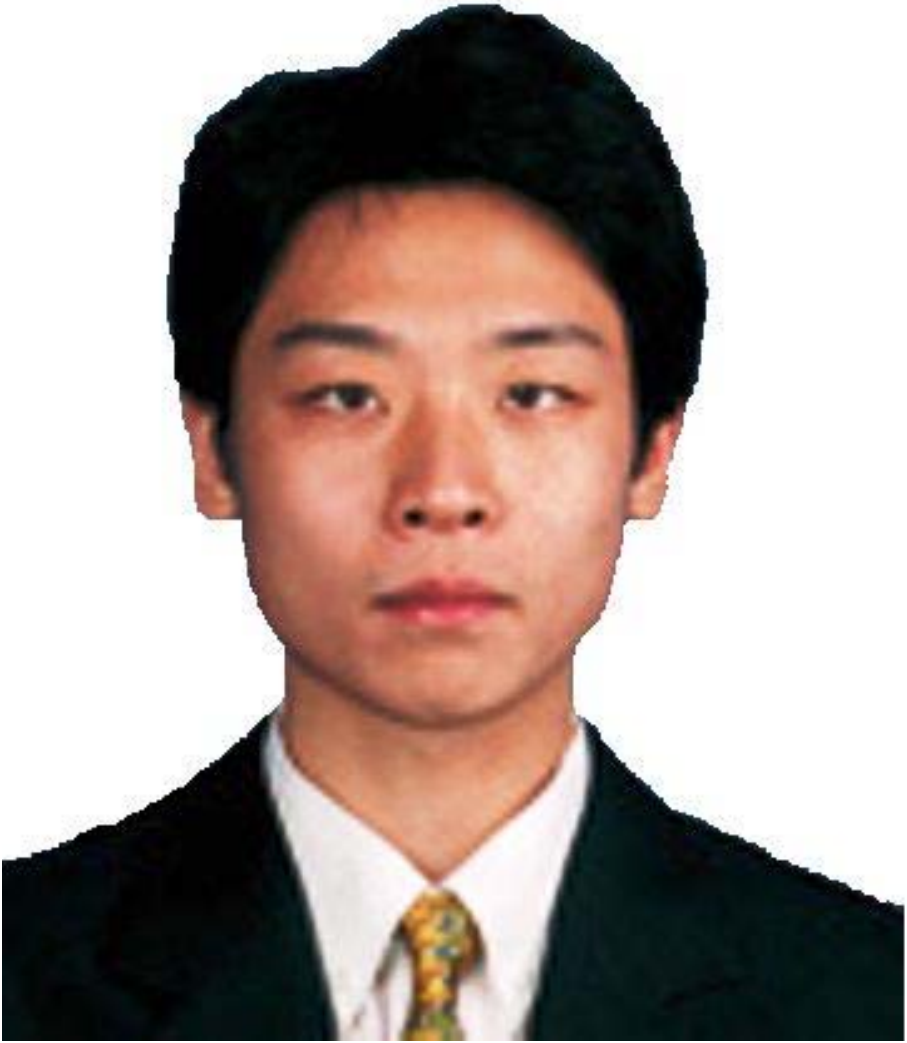}}]{Chi Harold Liu}(Senior Member, IEEE) receives a Ph.D. degree in Electronic Engineering from Imperial College, UK in 2010, and a B.Eng. degree in Electronic and Information Engineering from Tsinghua University, China in 2006.
He is currently a Full Professor and Vice Dean at the School of Computer Science and Technology, Beijing Institute of Technology, China. Before moving to academia, he worked for IBM Research, Beijing, China as a staff researcher and project manager from 2010 to 2013, worked as a postdoctoral researcher at Deutsche Telekom Laboratories, Berlin, Germany in 2010, and as a Research Staff Member at IBM T. J. Watson Research Center, USA in 2009. His current research interests include the big data analytics, mobile computing, and machine learning. He received the IBM First Plateau Invention Achievement Award in 2012, ACM SigKDD'21 Best Paper Runner-up Award, and IEEE DataCom'16 Best Paper Award. He has published more than 100 prestigious conference and journal papers and owned 39 EU/UK/US/Germany/Spain/China patents. He serves as the Associate Editor for IEEE TRANSACTIONS ON NETWORK SCIENCE AND ENGINEERING (with Excellent Editor Award, 2021), Area Editor for KSII Trans. on Internet and Information Systems, the Symposium Chair for IEEE ICC 2020 on Next Generation Networking, and served as the (Lead) Guest Editor for IEEE Transactions on Emerging Topics in Computing and IEEE Sensors Journal. He was the book editor for 11 books published by Taylor \& Francis Group, USA and China Machine Press, China. He also has served as the general chair of IEEE SECON'13 workshop on IoT Networking and Control, IEEE WCNC'12 workshop on IoT Enabling Technologies, and ACM UbiComp'11 Workshop on Networking and Object Memories for IoT. He was a consultant to Asian Development Bank, Bain \& Company, and KPMG, USA, and the peer reviewer for Qatar National Research Foundation, National Science Foundation, China, Ministry of Education and Ministry of Science and Technology, China. He is a senior member of IEEE and a Fellow of IET, British Computer Society, and Royal Society of Arts.
\end{IEEEbiography}


\begin{IEEEbiography}[{\includegraphics[width=1in,height=1.25in, clip, keepaspectratio]{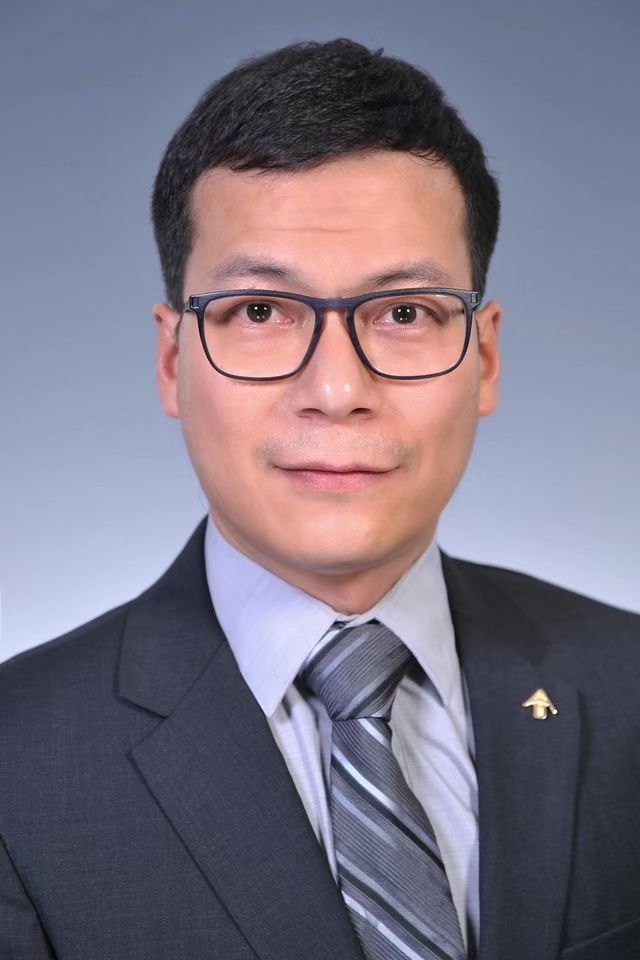}}]{Feifei Feng}
 received his B.S. and Ph.D. degrees in electronic engineering from Tsinghua University in 1999 and 2004, respectively. He is now with the AI Innovation Center, Midea Group.
His research interests include Internet of Things, artificial intelligence and ambient intelligence applications in smart home, and home service robotics.
\end{IEEEbiography}

\begin{IEEEbiography}[{\includegraphics[width=1in,height=1.25in, clip, keepaspectratio]{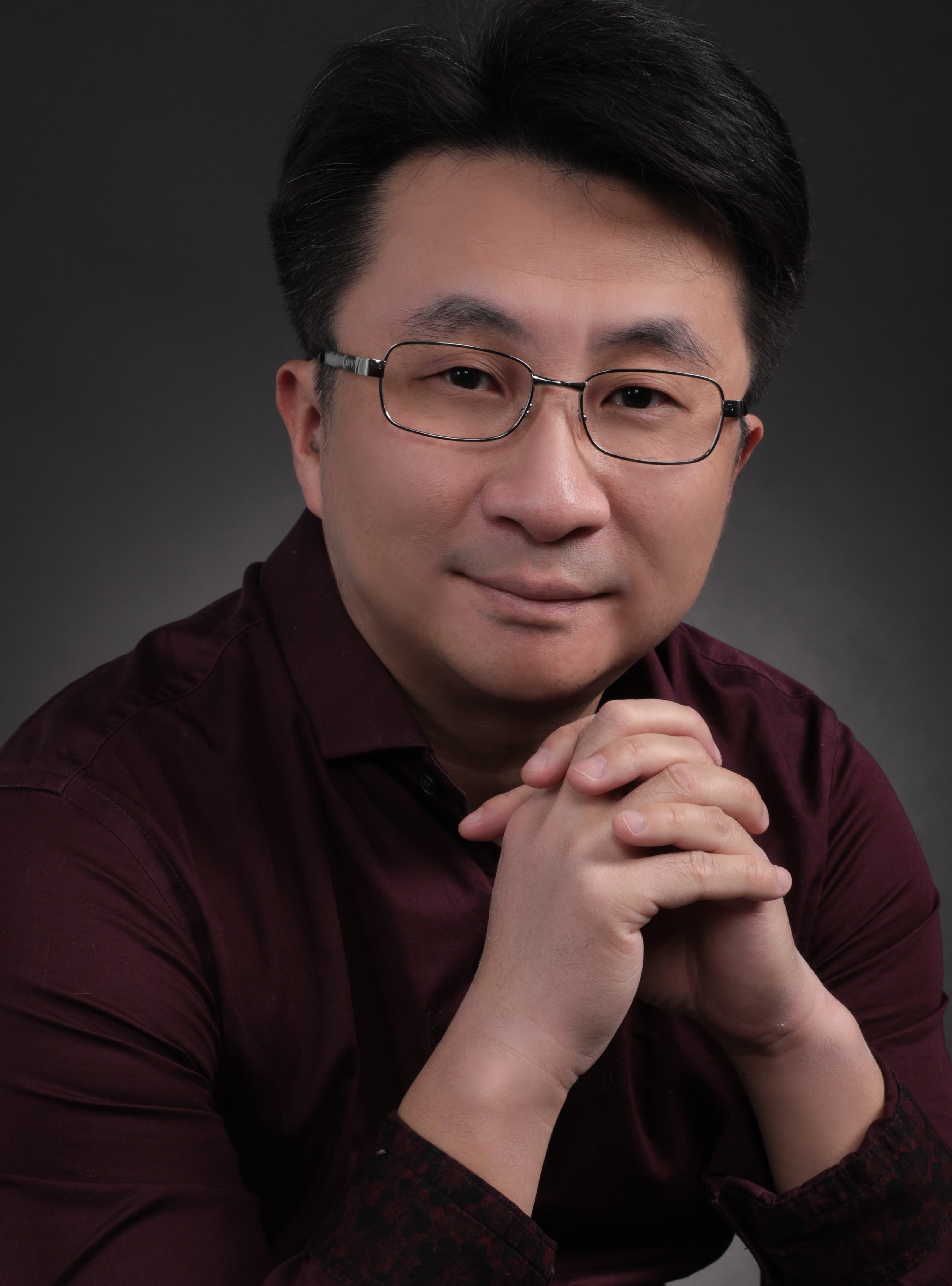}}]{Jian Tang}
(Fellow, IEEE) received his Ph.D. degree in Computer Science from Arizona State University in 2006. He is an IEEE Fellow and an ACM Distinguished Member. He is with Beijing Innovation Center of Humanoid Robotics, Beijing, China. His research interests lie in the areas of AI, IoT, and Robotics. He has published over 200 papers in premier journals and conferences. He received an NSF CAREER award in 2009. He also received several best paper awards, including the 2019 William R. Bennett Prize and the 2019 TCBD (Technical Committee on Big Data), Best Journal Paper Award from IEEE Communications Society (ComSoc), the 2016 Best Vehicular Electronics Paper Award from IEEE Vehicular Technology Society (VTS), and Best Paper Awards from the 2014 IEEE International Conference on Communications (ICC) and the 2015 IEEE Global Communications Conference (Globecom) respectively. He has served as an editor for several IEEE journals, including IEEE Transactions on Big Data, IEEE Transactions on Mobile Computing, etc. In addition, he served as a TPC co-chair for a few international conferences, including the IEEE/ACM IWQoS’2019, MobiQuitous’2018, IEEE iThings’2015. etc.; as the TPC vice chair for the INFOCOM’2019; and as an area TPC chair for INFOCOM 2017- 2018. He is also an IEEE VTS Distinguished Lecturer, and the Chair of the Communications Switching and Routing Committee of IEEE ComSoc 2020-2021.
\end{IEEEbiography}

\end{document}